\documentclass[final,5p,times,twocolumn]{elsarticle}

\usepackage{amssymb}
\usepackage{amsmath}
\usepackage[linesnumbered,ruled,vlined]{algorithm2e}
\usepackage{algpseudocode}

\usepackage{epsfig}
\usepackage{color,soul}
\setstcolor{yellow}
\usepackage{multirow}
\usepackage[table,xcdraw]{xcolor}
\usepackage[hyphenbreaks]{breakurl}
\usepackage{filecontents}
\usepackage{lipsum}
\usepackage{enumerate}
\usepackage[utf8]{inputenc}
\usepackage[T1]{fontenc}
\usepackage{comment}
\usepackage{indentfirst}
\usepackage{nomencl}
\renewcommand{\nomgroup}[1]{%
  \item[\textbf{%
    \ifthenelse{\equal{#1}{M}}{Parameters}{}%
    \ifthenelse{\equal{#1}{N}}{Variables}{}%
    \ifthenelse{\equal{#1}{I}}{Sets and Indices}
    }]%
}
\makenomenclature
\journal{Energy}

\begin{document}

\begin{frontmatter}

%% Title, authors and addresses

%% use the tnoteref command within \title for footnotes;
%% use the tnotetext command for theassociated footnote;
%% use the fnref command within \author or \address for footnotes;
%% use the fntext command for theassociated footnote;
%% use the corref command within \author for corresponding author footnotes;
%% use the cortext command for theassociated footnote;
%% use the ead command for the email address,
%% and the form \ead[url] for the home page:
%% \title{Title\tnoteref{label1}}
%% \tnotetext[label1]{}
%% \author{Name\corref{cor1}\fnref{label2}}
%% \ead{email address}
%% \ead[url]{home page}
%% \fntext[label2]{}
%% \cortext[cor1]{}
%% \address{Address\fnref{label3}}
%% \fntext[label3]{}

\title{Machine Learning Approach to Uncovering Residential Energy Consumption Patterns Based on Socioeconomic and Smart Meter Data}

%% use optional labels to link authors explicitly to addresses:
%% \author[label1,label2]{}
%% \address[label1]{}
%% \address[label2]{}

\author[A]{Wenjun~Tang}
\ead{monikatang@sz.tsinghua.edu.cn}
\author[B,C]{Hao~Wang\corref{mycorrespondingauthor}} 
\ead{hao.wang2@monash.edu}
\author[D]{Xian-Long~Lee}
\ead{xllee@ee.ncku.edu.tw}
\author[D]{Hong-Tzer~Yang\corref{mycorrespondingauthor}} 
\ead{htyang@ncku.edu.tw}

\cortext[mycorrespondingauthor]{Corresponding authors.}

\address[A]{Smart Grid \& Renewable Energy Lab, Tsinghua Berkeley Shenzhen Institute, Shenzhen 518055}
\address[B]{Department of Data Science and AI, Faculty of Information Technology, Monash University, Melbourne, VIC 3800, Australia}
\address[C]{Monash Energy Institute, Monash University, Melbourne, VIC 3800, Australia}
\address[D]{Department of Electrical Engineering, National Cheng Kung University, Tainan 70101} %Taiwan

\begin{comment}
\address{W. Tang is with Smart Grid \& Renewable Energy Lab, Tsinghua Berkeley Shenzhen Institute, Shenzhen 518055.
\protect\\
e-mail: monikatang@sz.tsinghua.edu.cn.
\protect\\
H. Wang is with the Department of Data Science and AI, Faculty of Information Technology and Monash Energy Institute, Monash University, Melbourne, VIC 3800, Australia. \protect\\
e-mail: haowang6@stanford.edu.
X. Lee and H. Yang are with the Department of Electrical Engineering, National Cheng Kung University, Tainan 70101. \protect\\
e-mails: xllee@ee.ncku.edu.tw; htyang@ncku.edu.tw.}
\end{comment}

\begin{abstract}
%% Text of abstract
The smart meter data analysis contributes to better planning and operations for the power system. This study aims to identify the drivers of residential energy consumption patterns from the socioeconomic perspective based on the consumption and demographic data using machine learning. We model consumption patterns by representative loads and reveal the relationship between load patterns and socioeconomic characteristics. Specifically, we analyze the real-world smart meter data and extract load patterns by clustering in a robust way. We further identify the influencing socioeconomic attributes on load patterns to improve our method's interpretability. The relationship between consumers' load patterns and selected socioeconomic features is characterized via 
machine learning models. The findings are as follows. (1) Twelve load clusters, consisting of six for weekdays and six for weekends, exhibit a diverse pattern of lifestyle and a difference between weekdays and weekends. (2) Among various socioeconomic features, age and education level are suggested to influence the load patterns. (3) Our proposed analytical model using feature selection and machine learning is proved to be more effective than XGBoost and conventional neural network model in mapping the relationship between load patterns and socioeconomic features.

%The further application can support the planning, pricing, and operation strategy making in the power system domain. \look{(False claim. No support can be found in this paper.)}
% Numerical studies validate our proposed framework using Pecan Street smart meter data and survey. We demonstrate that our framework can capture the relationship between load patterns and socioeconomic information and outperform benchmarks such as regression and single DNN models. 
\end{abstract}

%%Graphical abstract
% \begin{graphicalabstract}
% \includegraphics[width=1\columnwidth]{GA_.png}
% \end{graphicalabstract}

%%Research highlights
% \begin{highlights}
% \item The relationship between residential load and demographics is investigated.
% \item Residential energy consumption behavior is analyzed by the K-Medoid Algorithm.
% \item Feature selection is used to identify demographic features affecting load patterns.
% \item Deep learning is developed to reveal load patterns using socioeconomic information.
% \item Our method is validated using real-world smart meter data and demographic surveys.
% \end{highlights}

\begin{keyword}
%% keywords here, in the form: keyword \sep keyword
Consumption Pattern \sep Socioeconomic \sep Smart Meter \sep  Clustering\sep Feature Selection\sep Machine Learning
%% PACS codes here, in the form: \PACS code \sep code

%% MSC codes here, in the form: \MSC code \sep code
%% or \MSC[2008] code \sep code (2000 is the default)

\end{keyword}

\end{frontmatter}

%% \linenumbers
%\mbox{}

% \setlength{\nomlabelwidth}
\nomenclature[I]{$\mathcal{N}$}{Consumer set}%
\nomenclature[M]{$D$}{Total number of days}%
\nomenclature[I]{$\textit{w}$}{Index of weekday}%
\nomenclature[I]{$\textit{e}$}{Index of weekend}%
\nomenclature[I]{$t$}{Index of time}%
\nomenclature[I]{$d$}{Index of day}%
\nomenclature[N]{$l_{w,d,t}^n, l_{e,d,t}^n$}{Original load profile}%
\nomenclature[N]{$L_{w,d,t}^n, L_{e,d,t}^n$}{Normalized load profile}%
\nomenclature[N]{$\boldsymbol{L}_{w,d}^n, \boldsymbol{L}_{e,d}^n$}{Clustered load profile}%
\nomenclature[M]{$K$}{Number of the load cluster}%
\nomenclature[N]{${E}_{i,j}$}{Euclidean distance measurement}%
\nomenclature[N]{$\boldsymbol{\hat{L}}_w,\boldsymbol{\hat{L}}_e$}{Representative load profiles}%
\nomenclature[I]{$\boldsymbol{\hat{C}}_w,\boldsymbol{\hat{C}}_e$}{The set of profile indices in each cluster}%
\nomenclature[N]{$SC$}{Silhouette Coefficient index}%
\nomenclature[N]{$p_{k,n}$}{Load pattern distribution of consumer $n$}%
\nomenclature[I]{$\mathcal{S}$}{The set of socioeconomic features}%
\nomenclature[I]{$\mathcal{S}^k$}{The subset of socioeconomic features}%
\nomenclature[I]{$\mathcal{U},\mathcal{V}$}{Feature index}%
\nomenclature[N]{$\mu_n,v_n$}{Feature value}%
\nomenclature[N]{$P(\cdot)$}{Marginal probabilities}%
\nomenclature[N]{$H(\cdot)$}{Entropy}%
\nomenclature[N]{$MI(\cdot)$}{Mutual information}%
\nomenclature[N]{$SU(\cdot)$}{Symmetric uncertainty}%
\nomenclature[I]{$\mathcal{S}^k$}{The selected subset of feature(s)}%
\nomenclature[M]{$\boldsymbol{B}_l^k$}{The bias matrix of neural network}%
\nomenclature[N]{$\boldsymbol{o}_{l}^k$}{The output of layer $l$}%
\nomenclature[N]{$\sigma(\cdot)$}{The activation function of neural network}%
\nomenclature[N]{$\hat{p}_{k,n}$}{The intermediate-result of the prediction}%
\nomenclature[M]{$\boldsymbol{w}_l^k$}{The weight matrix of neural network}%
\printnomenclature

%% main text
\section{Introduction}
\label{sec:introduction}
Widespread deployment of smart meters generated a large volume of electricity consumption data. The newly available data of electricity consumption opened up opportunities for the utility to improve the system operation. %A comprehensive review about the analysis of smart meter data, e.g., forecasting, clustering, and classification is can be found in \cite{yildiz2017recent}.
%In addition to the algorithm designed for improving the performance of load profiling and load forecasting, 
Recent works took the perspective of the system operator or the utility to study how to utilize smart metre data analysis to enhance the system planning and operations through load forecasting \citep{barbato2011forecasting,candanedo2017data,shi2017deep}, demand response and customer behavior analysis \citep{wang2018review,lin2019clustering,sun2019clustering}, etc.

Smart meter data analysis provides insights into the electricity consumption patterns and characterizes consumption behaviors using load clustering \citep{ryu2019convolutional}. {From the review of \citep{wang2018review}, the main applications of smart meter data analytics can be classified into load analysis, load forecasting, and load management. The daily load curve can reflect the activities, e.g., cooking, cleaning, entertainment, and sleeping. The consumption behavior analysis can identify energy-saving potentials, plan energy supply, improve energy efficiency \citep{niu2021understanding}, and explore the diversity effect in residential energy consumption \citep{wang2020sizing}. It thus helps the system operator to determine the electricity tariff \citep{azarova2018exploring} and select consumers for various energy programs, e.g., demand response (DR) \citep{kwac2017efficient} and energy efficiency programs \citep{kwac2014household}.}
%in \cite{haben2015analysis}, a load forecasting strategy based on wavelet transform and artificial neural networks was proposed to predict the demand response of residential consumers to different price signals. 
A lifestyle segmentation method was developed in \cite{kwac2016lifestyle} to facilitate data-driven grid management using consumers' daily load shapes and consumption patterns. Utilizing the characterized consumption behavior, not only the energy management strategy but also the anomaly detection can be made \citep{capozzoli2018automated}.  Instead of focusing on the shape of the load curves, the work in \cite{wang2016clustering} investigated the transitions and relations between consumption behaviors based on the clustering results. Kwac and Rajagopal in \cite{kwac2015data} formulated a stochastic knapsack problem and utilized clustering results for customer selection in DR programs to minimize operational costs. More detailed literature review is presented in Section~\ref{sec:lit}.

From the discussions above, smart meter data can significantly improve the operation and consumer services of the utility through electricity consumption behavior analysis. Nevertheless, most related works rely on each consumer's historical load data to perform the consumption behavior analysis, which does not provide insights into the drivers of consumption behaviors. Studies in \citep{jones2015socio} revealed that energy consumption results from complex factors, such as socioeconomic and demographic factors. In addition, some households may not have been equipped with smart meters \cite{dang2015role}, or no historical data are available for new tenants. %In other words, system operators and utilities face a fundamental challenge that individual consumers' consumption behavior is often unknown and difficult to estimate, especially for the new house owners and tenants or when the load data are not available. 
A few recent studies \citep{rhodes2014clustering,tang2019isgt} took a new perspective %capasso1994bottom
to investigate the relationship between energy consumption and socioeconomic features, e.g., characteristics of the occupants. Meanwhile, the household natural gas consumption patterns and their influencing factors are explored using cluster analysis, taking into account the increasing block tariffs and temperature factors \citep{li2021exploring}. Thanks to the efforts made in the Pecan Street smart grid project, both smart meter data and household survey data become available \citep{street2016pecan}, enabling our study.

The above discussion motivates us to investigate the relationship between consumers' load patterns and socioeconomic features and improve the understanding of consumption behaviors.
In our work, using the real-world load and socioeconomic data, we reveal the electricity consumption patterns of consumers (e.g., using representative load curves), identify the main socioeconomic drivers of consumption patterns, and develop a framework to characterize the relationship between load patterns and socioeconomic features. To the best of our knowledge, our work is among the first to reveal the relationship between daily load patterns and socioeconomic factors using machine learning. We summarize the contributions of this paper as follows. 

\begin{itemize}
\item \textit{Load Pattern Extraction and Socioeconomic Feature Selection}: Given the advantage of the robustness to outliers and noises in real-world data, we model load patterns using the K-Medoids clustering, obtaining representative load profiles with associated probabilities.
%The Silhouette Coefficient is introduced to determine the number of typical load patterns.
We use an entropy-based feature selection algorithm to select the most correlated socioeconomic features with load patterns for enhancing the consumption behavior analysis and interpretability.

\item \textit{Tailored Model Using Deep Learning}: {We develop an analytical model by constructing pattern-dependent deep neural networks (DNNs) with a normalization layer to indicate the consumer's probability distribution of major load patterns only based on its socioeconomic information.}

\item \textit{Insights From Real-World Data}: We train and test our developed method based on real-world data.
    % of household load and the corresponding socioeconomic factors. 
    The results show strong correlations between the load patterns and selected socioeconomic features, improving the understanding of impact of socioeconomic factors on consumption behaviors.
    % of load patterns merely based on household socioeconomic information. 
    The evaluation results of load pattern distributions also demonstrate that our method outperforms benchmark methods, such as regression and unified DNN models. %, which can be employed to select the candidates for DR and energy efficiency program and moreover benefit the planning of energy tariff
\end{itemize}

The remainder of this paper is organized as follows. We review the related works in Section \ref{sec:lit}. We detail the analytical methodology, including clustering, feature selection, and deep learning model, in Section \ref{sec:methodology1}. 
% In Section \ref{sec:methodology2}, we employ an entropy-based algorithm to select the best subset of socioeconomic features and design a prediction model using DNNs to predict the probability distribution of load patterns. 
We discuss the simulation results in Section \ref{sec:simulations} and conclude this paper in Section \ref{sec:conclu}.

\section{Literature Review}
\label{sec:lit}

%Smart meter data have been recently used to enhance the operation and service of the utility \cite{wang2018review}. 
%A large body of literature in \citep{barbato2011forecasting,candanedo2017data,shi2017deep} has studied load forecasting, which benefits from the smart meter data. Barbato et al. \citep{barbato2011forecasting} developed a wireless power meter sensor network to collect consumption data of home devices and proposed an algorithm to predict the usage information of home devices. In addition to historical load data, \citep{candanedo2017data} considered the levels of load aggregation and weather conditions to improve the accuracy of short-term load forecast. Machine learning methods, especially deep learning methods, have been proven to be effective in the  short-term load forecast, e.g., in \citep{shi2017deep}. In \citep{wang2018ensemble}, load clustering was used to group individual load profiles to improve the load forecasting performance. %For a more comprehensive review on load forecast using smart meter readings, please refer to \cite{yildiz2017recent}. \look{(Not sure if forecasting should be included, as it is out of our scope. Load clustering is relevant.)}

% {Load management is an important issue 
% for electricity system stability and renewable energy application. Load clustering is a key topic of load management.
% } 
Load profiling emerges as a promising method to help the system operator and the utility understand the consumption behaviors \cite{kwac2016lifestyle} and enhance the operation and services, such as electricity tariff design \citep{azarova2018exploring}, consumer selection for demand response (DR) \citep{kwac2017efficient}, and energy efficiency programs \citep{kwac2014household}. The essential technique of load profiling is the clustering method, as clustering is an unsupervised learning method and has a great advantage in analyzing large-scale load datasets without labeled information. Specifically, clustering methods define the groups in an unsupervised way by organizing the data and placing similar objects into homogeneous groups \citep{ruhang2020efficient}. There are different clustering techniques \citep{motlagh2019clustering}, e.g., partitioning, hierarchical, grid-based, density-based, and feature-based methods, among which partitioning clustering is most widely employed to deal with time-series data. As one of the partitioning clustering methods, K-Means clustering is commonly used for shape-based load patterns due to its versatility on large datasets \citep{wang2018review}. 
% Besides the Euclidean distance, the cosine similarity, the discrete wavelet transform, and dynamic time warping distance are utilized as the similarity measure of the clustering method as well\cite{motlagh2019clustering}. 
However, the conventional K-Means method smooths out the temporal variations and may lose essential features \citep{scott2019clustering}. 
The K-Means method is also sensitive to the outliers and noises, resulting in narrow class margins among clusters and limitations in the interpretability of centroids \citep{park2009simple}. In contrast, the K-Medoids method can overcome the above drawbacks, and thus we use K-Medoids to process the real-world smart meter data.
% However, by representing the centroid as the average of the items in the cluster, K-Means reduce the time-to-time volatility\cite{scott2019clustering}.
% Additionally, some recent works used feature-based clustering and to classify the load profiles by describing the segmentation with the value of peak demand, the seasonal/annual average consumption, and other features derived from meter data \cite{rasanen2009feature,tong2016smart,williams2013clustering}. 
% Utilizing yearly\cite{ryu2019convolutional}, seasonal, and daily load profiles is vital for customer load clustering and analysis to account for long-term, mid-term and short-term characteristics of electricity load.  

{The socioeconomic information has been shown to contribute to the analysis of energy consumption \citep{rhodes2014clustering}, natural gas consumption \citep{li2021exploring}, and fuel poverty \citep{gouveia2018mining}. Through employing the cluster analysis, the work in \citep{li2021exploring} correlated the household natural gas consumption pattern with the socioeconomic factors, which aims to help improve energy efficiency and policy development. Daily smart meter data, combined with socioeconomic data, can enhance the understanding of energy consumption and the determinants of consuming behavior \citep{gouveia2018mining}.} Han et al. \citep{han2014impact} improved the forecasting performance of peak load and energy consumption by considering socioeconomic factors. The study in \citep{rhodes2014clustering} showed the relationship between the seasonal load patterns and the socioeconomic information. {And the structural and behavioral determinants of residential electricity daily peak and idle consumption can be estimated in a similar manner as well \citep{kavousian2013determinants}.} %In \citep{capasso1994bottom}, the authors revealed that energy consumption is correlated with socioeconomic factors, e.g., numbers of family members and their ages. 
%However, the black-box deep learning model makes the forecast results less explainable and thus often difficult to be deployed into a real application \citep{zhang2016understanding}. 
However, it lacks a systematic approach to understanding the effects of socioeconomic factors on energy consumption behaviors. 

Feature selection is proved to be a useful technique to remove the irrelevant and redundant features, improve efficiency and performance of learning tasks, and enhance the interpretability of the results \citep{yu2003feature}. Three primary types of methods are raised as feature selection algorithms, e.g., filter model, wrapper model, and embedded model \citep{witten2016data}.
%Wrapper models are in the way of searching the relevant set of features to leverage the characteristics of the specific classification algorithm to select features. Meanwhile, embedded models provide the solution to a classification model containing useful hints about the most related features. In other words, with the embedded models, the knowledge about the features is embedded in the solution to the classification. Compared with these two aforementioned models, 
The filter models can be employed to filter out irrelevant or redundant features \citep{bommert2020benchmark}. The filter model has higher computational efficiency compared with other feature selection models \citep{yu2003feature}.
% Specifically, 
% we employ the filter model \cite{witten2016data} for the feature selection. T
In our work, we use entropy-based class measurement \citep{jurado2015hybrid} to identify useful socioeconomic features affecting load patterns.
% Some mathematical criterions are available to estimate the quality of a feature or a subset of features in this model \cite{witten2016data}. 

%The authors in \cite{beckel2014revealing} estimated the household characteristics from its electricity consumption. 
In the smart grid project conducted by the Pecan Street team \citep{street2016pecan},
household socioeconomic data have been collected through energy audits together with the household consumption data. The collected socioeconomic data enable us to study how the energy customers' socioeconomic characteristics drive the consumption patterns.
Recent works in \citep{rhodes2014clustering,tang2019isgt,viegas2016classification} attempted to study the relationship between energy consumption and socioeconomic status, and the preliminary results showed that the peak load and energy consumption are correlated with socioeconomic factors. However, how socioeconomic factors affect residential energy consumption patterns is not well understood. This motivates us to develop a systematic framework to reveal the relationship between load patterns and socioeconomic factors using machine learning.
% This motivates our study to estimate load patterns using socioeconomic data when meter data are not available. Our preliminary results in \cite{tang2019isgt} revealed the relationship between load patterns and socioeconomic information using K-Means clustering and deep learning.

% Irish Social Science Data Archive \cite{viegas2016classification}, 

% In our work, we aim to take a further step to characterize consumers' load patterns, identify the correlations of their socioeconomic background with load patterns, and eventually help system operators predict consumer behaviors even without knowing their historical electricity consumption.

% %\vspace{-0.4cm}

\section{Load Patterns and Relationship with Socioeconomic Factors}
\label{sec:methodology1}
\begin{figure}[!tbp] \centering
\includegraphics[width=0.9\columnwidth]{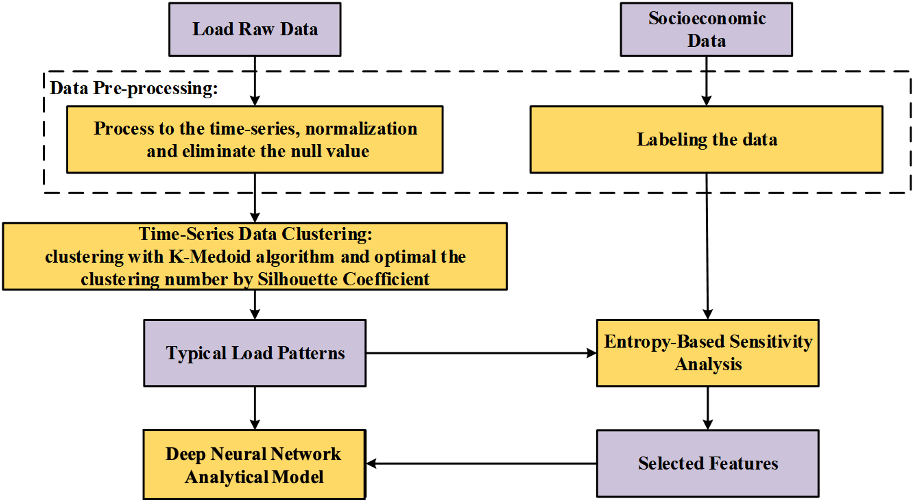}
\caption{The flowchart of our study on the relationship between load patterns and socioeconomic factors.} 
\label{fig1} 
\end{figure}

In this section, we present our method to uncover the relationship between load patterns and socioeconomic factors. Figure \ref{fig1} describes the flowchart of the proposed framework, and the details are presented in the following subsections.

\begin{itemize}
    \item In Subsection \ref{sec:processing}, the consumption data are normalized to capture the temporal variations of daily time-series load, and the socioeconomic data are also processed.
    \item In Subsection \ref{sec:clustering}, the typical load patterns are characterized by the K-Medoids clustering method, and the number of clusters is determined using the Silhouette Coefficient index.
    \item In Subsection \ref{feature_sec}, the socioeconomic features are selected based on their correlations with the clustered load patterns.
    \item In Subsection \ref{subsec:prediction}, we build a tailored analytical model with pattern-dependent deep neural networks and a normalization layer to study the relationship between load patterns and selected socioeconomic features.
\end{itemize}

%  data pre-processing for load and socioeconomic data.
% % we use residential load data from Pecan Street \citep{street2016pecan} to study the consumer load patterns. 
% We then leverage K-Medoids method to characterize consumer load patterns on weekdays and weekends, respectively.  
% and K-Medoids clustering method in the following.
% \vspace{-0.4cm}

\subsection{Load Data Pre-processing and Socioeconomic Labels}\label{sec:processing}
{Since the daily energy consumption is different on weekdays and weekends \citep{zhang2008grouping,wang2015load}, we divide the dataset into two groups for the workday and weekend per the date stamps. 
% Due to the difference in consumption behaviors on weekdays and weekends, 
Specifically, we divide the load data of all the consumers (denoted as $\mathcal{N} = \{1,...,N \}$) into two parts: weekdays (denoted as \textit{w}) and weekends (denoted as \textit{e}).} We denote $D$ as the total number of days in the load data and each day is denoted as $d \in \{1,...,D\}$. Note that there are $D_w$ days and $D_e$ days in total for weekdays and weekends, respectively. Each day $d$ is divided into 24 one-hour intervals, i.e., $t =1,..., 24$. The load profile of consumer $n \in \mathcal{N}$ in hour $t$ on day $d$ is defined as $l_{w,d,t}^n$ and $l_{e,d,t}^n$ separately for weekdays and weekends. As we focus on the temporal variations of load, we normalize the original load profiles $l_{w,d,t}^n$ and $l_{e,d,t}^n$ to be normalized ones $L_{w,d,t}^n$ and $L_{e,d,t}^n$ in the range of $[0,1]$ shown as
\begin{equation}
    L_{w,d,t}^n = \frac{l_{w,d,t}^n - \min_t \{l_{w,d,t}^n\}}{\max_t \{l_{w,d,t}^n\} - \min_t \{l_{w,d,t}^n\}},
\end{equation}
% \begin{equation}
% L_{w,d,t}^n=\frac{l_{w,d,t}^n-\min_t ⁡\{l_{w,d,t}^n\}}{\max_t⁡ \{l_{w,d,t}^n\}-\min_t⁡ \{l_{w,d,t}^n\}},
% \end{equation}
% \begin{equation}
% L_{w,d,t}^n=\frac{l_{w,d,t}^n-min_t ⁡\{l_{w,d,t}^n\}}{max_t⁡ \{l_{w,d,t}^n\}-min_t⁡ \{l_{w,d,t}^n\}},
% \end{equation}
and
\begin{equation}
    L_{e,d,t}^n = \frac{l_{e,d,t}^n-\min_t \{l_{e,d,t}^n\}}{\max_t \{l_{e,d,t}^n\} - \min_t \{l_{e,d,t}^n\}},
\end{equation}
% \begin{equation}
% {
% L_{e,d,t}^n=\frac{l_{e,d,t}^n-\min_t ⁡\{l_{e,d,t}^n\}}{\max_t⁡ \{l_{e,d,t}^n\} -\min_t⁡ \{l_{e,d,t}^n\}},}
% \end{equation}
where $\max_t \{l_{w,d,t}^n\}$ and $\min_t \{l_{w,d,t}^n\}$ (or $\max_t \{l_{e,d,t}^n\}$
and $\min_t \{l_{e,d,t}^n\}$) denote the maximum and minimum hourly load of consumer $n$ on weekdays $w$ (or weekends $e$) of day $d$.

% The incomplete daily profiles with null values are ignored as the invalidity to our following analysis. \look{We haven't talked about real data in this part and thus there is no problem about null values.}

For socioeconomic factors, it consists of different attributes, namely features, with various data types, shown as follows.
\begin{itemize}
    \item \textit{The number of residents in six different age ranges} (\textit{integer variable});
    \item \textit{Annual income} with ten ranges (\textit{categorical variable});
    \item \textit{Education level} with four categories (\textit{categorical variable}); 
    \item \textit{Total square footage} of the residents (\textit{integer variable}).
\end{itemize}
The ordered scales of the categorical variable are extracted into a matrix of metadata and expressed as the quantified ordinal classification. For the \textit{annual income} and \textit{education level}, the categories are labeled, as shown in Table \ref{label}.

\begin{table}[!btph]
\centering
\caption{The Quantified Ordinal Classification of The Features}
% \vspace{-0.3cm}
\label{label}
\resizebox{\columnwidth}{!}{
\begin{tabular}{|c|c|c|c|}
\hline
\textit{Annual Income Range} & Value& \textit{Education Level} & Value\\ \hline \hline
Less than \$10,000 & 1 & \multirow{2}{*}{High School graduate} & \multirow{2}{*}{1} \\ \cline{1-2}
\$10,000 - 19,999 & 2 &  &  \\ \hline
\$20,000 - 34,999 & 3 & \multirow{3}{*}{\begin{tabular}[c]{@{}c@{}}Some college/trade\\ /vocational school\end{tabular}} & \multirow{3}{*}{2} \\ \cline{1-2}
\$35,000 - 49,999 & 4 &  &  \\ \cline{1-2}
\$50,000 - 74,999 & 5 &  &  \\ \hline
\$75,000 - 99,999 & 6 & \multirow{2}{*}{College graduate} & \multirow{2}{*}{3} \\ \cline{1-2}
\$100,000 - 149,999 & 7 &  &  \\ \hline
\$150,000 - 299,000 & 8 & \multirow{2}{*}{Postgraduate degree} & \multirow{2}{*}{4} \\ \cline{1-2}
\$300,000 - 1,000,000 & 9 &  &  \\ \hline
\end{tabular}}
\end{table}

\subsection{Load Pattern Clustering by K-Medoids Algorithm} \label{sec:clustering}
% As one of the clustering algorithms, K-Means method is widely used due to its simplicity, efficiency, and interpretable characteristics \citep{wagstaff2001constrained}. 
% K-Means clustering algorithm partitions data samples into $K$ clusters, and finds the centroid to represent each cluster. Clustering the data samples is based on some distance measures between the sample data and centroids, and Euclidean distance is a popular measure. % The centroid of the cluster is then calculated by the means of the objects in a certain cluster.
Real-world data samples often contain irregular data and outliers. 
As discussed, the conventional K-Means method is sensitive to outliers and noises, and the centroids can be greatly affected by outliers.
%, becoming not representative.
%For example, if the data valued $1,~2,~3$, and an outlier $100$ are clustered into the same group, the centroid is calculated as $26.5$, which is obviously not representative. 
% The drawbacks of the K-Means algorithm imposed by challenging real-world data motivated the new design of clustering algorithms. 
% K-Medoids algorithm was proposed to overcome the drawbacks of K-Means algorithm \citep{park2009simple}. 
Instead of finding centroids by the average, the K-Medoids method searches for the most central and representative sample in a cluster. It is also more robust to outliers and noises, compared with the K-Means method. Moreover, in our case, the K-Medoids algorithm derives clusters using actual load profiles rather than the mean of profiles and thus does not smooth out the temporal variations. 
% retake the aforementioned example, we see that the data valued $3$ is computed as the medoid of the group using the K-Medoids, which weakens the impact of outliers.

We use the K-Medoids method to characterize consumer load patterns, specifically by clustering the daily load profiles $\boldsymbol{L}_{w,d}^n={(L_{w,d,t}^n,~t=1,…,24)}$ and $\boldsymbol{L}_{e,d}^n={(L_{e,d,t}^n,~t=1,…,24)}$ of all consumers $n \in \mathcal{N}$ for weekdays and weekends, respectively. Without loss of generality, we assume that the K-Medoids algorithm is going to determine $K$ clusters as load patterns on both weekdays and weekends.\footnote{We will discuss how to select the number of clusters $K$ at the later part from Eq. \eqref{cohesion} of Section \ref{sec:methodology1}.} In the following, we take the clustering on weekdays as an example and use Euclidean distance measure for each pair of load profiles $i,j \in \{1,..., N \times D_w\}$ as 
\begin{equation}
{E}_{i,j} = \lVert s_{i} - s_{j} \rVert,
\label{dist}
\end{equation}
where $s_{i},~s_{j} \in \boldsymbol{L}_{w,d}^n$, and $d =1,...,D_w$. 

Based on the distance calculation, we use the K-Medoids algorithm in Algorithm \ref{alg:kmedoid} to obtain $K$ representative load profiles $\boldsymbol{\hat{L}}_w=\{\hat{L}_w^1,\dots,\hat{L}_w^K\}$ and a set of profile indices in each cluster denoted as $\boldsymbol{\hat{C}}_w=\{\boldsymbol{\hat{C}}_w^1,\dots,{\boldsymbol{\hat{C}}}_w^K\}$. 
% In the algorithm, we define the dissimilarity of the medoids and the load profiles
% \begin{equation}
%     c=\sum_
% \end{equation}
% as the cost of K-Medoids algorithm 
Specifically, we randomly select $K$ load profiles from $\boldsymbol{L}_{w,d}^n$ as the initial medoids. All the load profiles in $\boldsymbol{L}_{w,d}^n$ are then separately calculated with the medoids based on the distance measure. For the initial $K$ medoids, the shortest distance between the profile and the medoid determines the allocation of each profile in the nearest cluster as described in step 4. In step 5, the cluster score is defined as the summation of the total distance between each profile to its medoid. Afterward, the algorithm proceeds iteratively from step 6 to 12. Each iteration begins with the pairwise distance measurements among the profiles in the same cluster. The profile, having the overall shortest distance with the rest of others in the same cluster, is updated to be the new medoid. 

We update the cluster referring to the new medoid and repeat all the steps to assign all load profiles in the corresponding clusters based on the shortest-distance rule. At each iteration denoted as $iter$, $\boldsymbol{L}_w^{(iter)}$ and $\boldsymbol{C}_w^{(iter)}$ are employed to store the temporary results for $\boldsymbol{\hat{L}}_w$ and $\boldsymbol{\hat{C}}_w$. The iterations end once the cluster score does not decrease. The algorithm outputs the medoids in $\boldsymbol{\hat{L}}_w$ representing the major load patterns on weekday and $\boldsymbol{\hat{C}}_w$ containing the profile indices corresponding to each pattern.
% proceeds with the steps shown in Algorithm \ref{alg:kmedoid}. The representative load profiles $\boldsymbol{\hat{L}}_w=[\hat{L}_w^1,\dots,\hat{L}_w^K]$ can thus be obtained to represent \textit{K} different patterns on weekdays. Meanwhile, the load profiles in $\boldsymbol{L}_{w,d}^n$ are clustered into the $K$ clusters with their index, e.g. $i$ and $j$, be put into the set $C_k$.
% $\boldsymbol{{L}}_w=[\boldsymbol{\hat{L}}_{w,d}^n,d=1,\dots,D_w,n=1,\dots,N]$. 
Similarly, we can obtain the weekend load patterns $\boldsymbol{\hat{L}}_e$ and the index set $\boldsymbol{\hat{C}}^k_e$ using Algorithm \ref{alg:kmedoid} based on the weekend load profiles $\boldsymbol{L}_{e,d}^n$. 
%\rewrite{$\{ \boldsymbol{L}_{e,d}^n,~d=\{1,...,D_e\},n\in \mathcal{N}\}$}.
\begin{algorithm}[t!]
\caption{Load Clustering Algorithm.}
\label{alg:kmedoid}
\LinesNumbered
 \KwIn{The number of clusters \textit{K} and the total daily load profiles $\boldsymbol{L}_{w,d}^n$}
 \KwOut{The \textit{K} representative load pattern $\boldsymbol{\hat{L}}_w$ and index set $\boldsymbol{\hat{C}}_w$}
 Iteration time $iter$ =0\;
 Randomly select $K$ load profiles from $\boldsymbol{L}_{w,d}^n$ as the initial medoids  $\boldsymbol{{L}}_w^{(iter)}$  \;
 For each profile $i$, determine its belonging $\boldsymbol{C}^{(iter)}_w$ per the shortest distance to the medoids, i.e.,
 $\mathop{\arg\min}_{k \in \{1,\dots,K\}} \lVert s_{i} - {L}_w^{k,(iter)} \rVert,$ where $s_i \in \boldsymbol{L}_{w,d}^n$\;
 and $s_j \in \boldsymbol{{L}}_w^{(iter)}=\{\hat{L}_w^1,\dots,\hat{L}_w^K\}$\;
 $Cluster Score^{(iter)}=\sum^K_{k=1}\sum_{i \in \boldsymbol{C}_w^{k,(iter)}}\lVert s_{i} - {L}_w^{k,(iter)} \rVert$\;
 \For{Iteration time $iter \leq iter^{max}$}{$iter=iter+1$\;Update ${L}_w^{k,(iter)}$ with the profile whose index in $\boldsymbol{C}_w^{k,(iter-1)}$ through $\arg\min_i \sum_{i,j \in \boldsymbol{C}_w^{k,(iter-1)}} E_{i,j}$\;Update $\boldsymbol{C}_w^{k,(iter)}$  refer to \textit{Step 3} \;Update  $Cluster Score^{(iter)}$\ refer to \textit{Step 4} \;\If{ $Cluster Score^{(iter)}=Cluster Score^{(iter-1)}$}{break\;}}
 $\boldsymbol{\hat{L}}_w=\boldsymbol{{L}}_w^{(iter)}$,
 $\boldsymbol{\hat{C}}_w=\boldsymbol{C}^{(iter)}_w$
%  \For{Iteration time $iter \leq iter^{max}$}{
% 　　$iter=iter+1$\;
% 　　Update ${L}_w^{k,(iter)}$ with the profile whose index in $\boldsymbol{C}_w^{k,(iter-1)}$ through $\arg\min_i \sum_{i,j \in \boldsymbol{C}_w^{k,(iter-1)}} E_{i,j}$\;
%       Update $\boldsymbol{C}_w^{k,(iter)}$  refer to \textit{Step 3} \;
%     %   where $\boldsymbol{L}_w^{(iter)}$ is used instead of $\boldsymbol{L}_w^{(0)}$\;
%       Update  $Cluster Score^{(iter)}$\ refer to \textit{Step 4} \;
% 　　\If{ $Cluster Score^{(iter)}=Cluster Score^{(iter-1)}$}{
% 　　　　break\;
% 　　}
% }
% $\boldsymbol{\hat{L}}_w=\boldsymbol{{L}}_w^{(iter)}$,
% $\boldsymbol{\hat{C}}_w=\boldsymbol{C}^{(iter)}_w$
\end{algorithm}

Despite the advantage of robustness, there still exists a problem for K-Medoids clustering being widely discussed, e.g., a heavy computational overhead. As defined in \eqref{dist}, the pairwise calculation is required in K-Medoids clustering, which directly causes high overhead with a large number of data samples. Therefore, we adopt an improved K-Medoids algorithm, $trimed$, \citep{newling2017sub} to enhance the computational efficiency. The improved algorithm $trimed$ can reduce the computational complexity from $O(N^{2})$ to $O(N^{3/2})$.
%by using parallel computing and setting boundaries to avoid recomputation.

We further investigate how to determine the value of \textit{K} using the Silhouette Coefficient index \citep{zhu2010clustering}. Specifically, the Silhouette Coefficient index $SC$ consists of two parts: the cohesion factor $a_i$ and the separation factor $b_i$, which are defined as 
\begin{equation}\label{cohesion}
a_i=\frac{\sum_{j\in \boldsymbol{\hat{C}}_w^{k} \backslash i} E_{i,j}}{M_w^k-1},    
\end{equation}
\begin{equation}
b_i=\min_{ k'\in \{1,..,K\}\backslash k}\frac{\sum_{j\in \boldsymbol{\hat{C}}_w^{k'}} E_{i,j}}{M_w^{k'}}   
\end{equation}
where $k =1,...,K$, and $M_w^k$ represents the number of load profile indices in $\boldsymbol{\hat{C}}_w^{k}$. Factor $a_i$ stands for the mean distance between the \textit{i}-th load profile and other load profiles in the same cluster $k$. Factor $b_i$ measures \textit{i}-th load profile's minimum averaged distance to load profiles of other load clusters $k'\in \{1,..,K\}\backslash k$. 

Then we obtain $SC_i$ for load profile $i$ as
\begin{equation}
    SC_i=\frac{b_i-a_i}{\max(b_i,a_i)},
\end{equation}
which is employed to estimate whether the assignment of the \textit{i}-th load profile is appropriate. A smaller cohesion factor $a_i$ and a larger separation factor $b_i$ are preferred. 

The overall clustering performance is thus measured by $SC$ as
\begin{equation}
    SC=\frac{\sum_{i=1}^{M_w}SC_i}{M_w},
\label{SCeq}\end{equation}
which is ranged within $[-1,1]$ and
% . We let
% $M_w$ denote 
% the total number of load profiles on weekday data, i.e., 
$M_w \triangleq N \times D_w$. A higher Silhouette Coefficient index determines a better descriptive number of \textit{K}. 
% \look{what is $M$? It's not defined. If it's the total number of load profile, then it should be defined as $M \triangleq N \times D$.}

After obtaining the major load patterns, we further calculate the load pattern distribution of consumer $n$ in pattern $k$ by
% . We count up the number of load profiles of consumer $n$ in the $k$-th cluster denoted as $M_{k,n}$ and the corresponding percentage is
\begin{equation}
    p_{k,n}=\frac{M_{k,n}}{\sum_{k=1}^{K}M_{k,n}},
\end{equation}
where the number of load profiles of consumer $n$ in the $k$-th cluster denoted as $M_{k,n}$. We will use $p_{k,n}$ together with consumers' socioeconomic factors to further study their correlations.
% \vspace{-0.5cm}

% \vspace{-0.5cm}

\subsection{Load Pattern-Related Socioeconomic Features Selection}
\label{feature_sec}

Real-world data often contain irrelevant and redundant features as well and thus can cause low accuracy, unnecessarily complex modeling, and high computational burden for prediction models. Feature selection plays a key role in eliminating redundant features and selecting the best subset of predictors (i.e., features) \citep{zhu2010clustering}. With the determined load patterns in Subsection \ref{sec:clustering}, we explore the correlations between the socioeconomic features and the load patterns.
% represented by load patterns. 

In this subsection, we present how to select socioeconomic features via an entropy-based filter method. In our problem, the socioeconomic features include different types, e.g., \textit{age}, \textit{education}, \textit{income}, and \textit{household square footage}, but how these features are correlated with each load pattern remains unknown.
% To select the most predictive features respecting the class label, we raise filter model as the tool for feature selection. 
% three primary types of methods are raised as feature selection algorithms, e.g. filter model, wrapper model, and embedded model \citep{witten2016data}.
% Wrapper models are in the way of searching the relevant set of features to leverage the characteristics of the specific classification algorithm to select features. Meanwhile, Embedded models provide the solution to a classification model that contains useful hints about the most related features. In other words, with the embedded models, knowledge about the features is embedded in the solution to the classification. Compared with these two afore-mention models, 
% Literally as defined to be ‘filter’, the filter models are employed to filter out irrelevant or redundant features. Some mathematical criterions are available to estimate the quality of a feature or a subset of features in this model \citep{witten2016data}. 
% Conclusively, 
Therefore, we use feature selection techniques to identify the relevant features for the prediction of the distribution of load patterns. Specifically, 
% we employ the filter model \citep{witten2016data} for the feature selection. T
the entropy-based class measurement \citep{jurado2015hybrid}, which has been widely employed in filter models \citep{witten2016data}, is used in this paper.
% \look{Among [31-33] we can only use one reference here. It's only one method, just keep one reference. If we combine different parts from [31-33], we cannot say `the filter model' but a xxx method using xxx [31], xxx[32], and xxx [33]. In this case, you need put more words to present our model. If it's not the case, just choose the one reference that we use the most.}

We define the set of features, including \textit{the number of residents in each age range}, \textit{annual income}, \textit{educational level}, and \textit{the total foot square}, as $\mathcal{S}$. We aim to select a subset of features $\mathcal{S}^k \subseteq \mathcal{S}$ for each load pattern \textit{k} to minimize the redundancy. To determine the best subset $\mathcal{S}^k$, we evaluate all combinations of features $\mathcal{S}' \subseteq \mathcal{S}$.
% based on the entropy-based methodology \citep{jurado2015hybrid}. 
The entropy-based methodology measures the intercorrelation among the features in $\mathcal{S}'$ pairwisely and the correlations between features in $\mathcal{S}'$ and predicted target, i.e., the percentage/probability of the user's load profiles in pattern cluster $k$ denoted by $\boldsymbol{p}_{k}=(p_{k,n},n=1,\dots,N)$.
% \look{add the reference for it.}} 
Then the subset with the highest measurement value is determined to be $\mathcal{S}^k$, which selects the features having a high correlation with the evaluating target. 

We denote $\mathcal{U}$ and $\mathcal{V}$ as two different features in $\mathcal{S}'$. For example, the features \textit{number of resident age under 13} and \textit{education level} can form the feature subset $\mathcal{S}'$.
% , features can be the . 
We let $\mu_n \in \mathcal{U}$ and $v_n \in \mathcal{V}$ denote randomly selected feature values for consumer $n \in \mathcal{N}$ from $\mathcal{U}$ and $\mathcal{V}$, respectively. For the \textit{number of resident age under 13}, the value of $\mu_n$ can be `0', `1', or other integers. If the underlying feature is the \textit{education level}, the corresponding values are `1', `2', `3', and `4' representing four education levels from the high school to graduate school.

We denote $P(\mu_n)$ and $P(v_n)$ as the marginal probabilities for the feature values. The marginal probability $P(\mu_n=0)$ of the feature \textit{number of resident age under 13} therefore represents the statistical probability of value `0' among all the possible realizations for the underlying feature. Similar to the marginal probability $P(\mu_n)$, the joint probability $P(\mu_n,v_n)$,
% , with the similar idea of marginal probability $P(\mu_n)$, 
represents the statistical probability of the value $\mu_n$ and $v_n$ appearing at the same time in $\mathcal{U}$ and $\mathcal{V}$.
% with marginal probabilities $P(\mu)$ and $P(v)$, respectively. 

The entropy $H(\mathcal{U})$, as a measure of distribution property, is calculated based on $P(\mu_n)$ by 
% \look{We must make the definition of $\mathcal{U}$ consistent. What is $\mathcal{U}$? You write feature space for it. But you say it's a feature category later.}
\begin{equation}
H(\mathcal{U}) = -\sum_{\mu_n \in \mathcal{U}}P(\mu_n)\log{P(\mu_n)},
\end{equation}
where $H(\mathcal{U})$ ranges from 0 to 1. % \look{To help readers understand $\mathcal{U}$, we should add a couple sentences to give an example. Choose socioeconomic features in our paper to illustrate the definitions of $\mathcal{S’}$, $\mathcal{U}$, and $\mu$.}
The higher level of disorder of data in $\mathcal{U}$ corresponds to a greater value of $H(\mathcal{U})$.  
Based on the idea of entropy evaluation and the joint probability, mutual information $MI(\mathcal{U},\mathcal{V})$ is a technique to measure how much knowledge between two features is correlated. It is defined as the difference between the sum of the marginal entropy and their joint entropy \citep{sarhrouni2012application} and written as 
% Based on the idea of entropy evaluation, mutual  information $MI(\mathcal{U},\mathcal{V})$ \citep{sarhrouni2012application}, is obtained to measure the dependency of the two features} \look{Why we need MI?}  \look{dependency of what? what to do with our socioeconomic features?} by
\begin{equation}
MI(\mathcal{U},\mathcal{V})=\sum_{\mu_n \in \mathcal{U}}\sum_{v_n \in \mathcal{V}}P(\mu_n,v_n)\log{\frac{P(\mu_n,v_n)}{P(\mu_n) P(v_n)}}.    
\end{equation}

Note that dependency $MI(\mathcal{U},\mathcal{V})$ reflects the correlation of the features and
is always nonnegative. It is zero if and only if $\mathcal{U}$ and $\mathcal{V}$ are independent. A stronger dependency between $\mathcal{U}$ and $\mathcal{V}$ is revealed when $MI(\mathcal{U},\mathcal{V})$ is relatively large.

The symmetric uncertainty ($SU$) is one of the normalized forms of $MI$ and is defined as 
% The symmetric uncertainty ($SU$) is introduced, which normalizing  $MI(\mathcal{U},\mathcal{V})$ to the individual entropy value of $\mathcal{U}$ and $\mathcal{V}$, i.e.,
% \look{Why we need SU?} 
\begin{equation}
SU(\mathcal{U},\mathcal{V})=2 \frac{MI(\mathcal{U},\mathcal{V})}{H(\mathcal{U})+H(\mathcal{V})},    
\end{equation}
where $SU$ ranges in [0,1].
% $SU$ measures how much shared information is distributed at all information contained in both $\mathcal{U}$ and $\mathcal{V}$ (e.g., $MI(\mathcal{U},\mathcal{V})$) by the standard set up by the individual information distribution of $\mathcal{U}$ and $\mathcal{V}$ (e.g. $H(\mathcal{U})$ and $H(\mathcal{V})$).
% \textcolor{olive}{$SU$ ranges between [0,1], measuring the inner correlation between features. 
The value `1' indicates the knowledge of $\mathcal{U}$ completely predicts that of $\mathcal{V}$ where the value `0' indicating that $\mathcal{U}$ and $\mathcal{V}$ are independent.
Similarly, the target-related $SU$, e.g., $SU(\mathcal{U},\boldsymbol{p}_{k})$, is calculated to estimate the correlation between the feature $\mathcal{U}$ and target $\boldsymbol{p}_{k}$. 
% The two spaces, the same, are linked by the consumer index '$n$'. 

Finally, based on the calculated symmetric uncertainty, we select the subset of features $\mathcal{S}^k$ as 
\begin{equation}
% \label{fea_eq}
 \mathcal{S}^k=\mathop{\arg \max}_{\mathcal{S'} \subseteq \mathcal{S}} \frac{\sum_{\mathcal{U}\in S'}SU(\mathcal{U},\boldsymbol{p}_{k})}{\sqrt{\sum_{\mathcal{U} \in S'}\sum_{\mathcal{V} \in S'}SU(\mathcal{U},\mathcal{V}) }},
 \label{featuresele.equ}
\end{equation}
% \look{Here, $p_{k,n}$ is not accurate, as it contains index $n$ but there is no $n$ in $\mathcal{U}$. They are not at the same dimension. If you did the numerical simulation/calculation, you should know precisely what is $u$, $\mathcal{U}$, and $p_{k,n}$. Should $p_{k,n}$ be $\boldsymbol{p}_{k}$? If so, we also need to define $\boldsymbol{p}_{k}$ and use example to illustrate its meaning.}
where $\mathcal{S}^k$ is determined to select the feature(s) that are highly correlated with the target but less correlated with each other. For each load pattern $k$, we run Equation (\ref{featuresele.equ}) over all combinations of feature $\mathcal{S'} \subseteq \mathcal{S}$ to obtain $\mathcal{S}^k$. The $K$ selected feature subsets are finally obtained as $\{\mathcal{S}^k,~k=1,\dots,K\}$ and will be used as the indicators in the 
analytical model in Section \ref{subsec:prediction}.

% \look{too long, not clear. Break into two sentences.}  \look{Before introducing Algorithm 2, what does `algorithm' refer to?} \look{We need to introduce Algorithm 2.} 

% \begin{algorithm}[tp!]
%   \textcolor{olive}{
%   \KwIn{The set of socioeconomic features $\mathcal{S}$ and the target spaces $\{\boldsymbol{p}_{k},k=1,\dots,K\}$.}
%   \KwOut{The subset of selected socioeconomic features for \textit{k}-th load pattern $\{\mathcal{S}^k,k=1,\dots,K\}.$}
% Construct the subset $\mathcal{S}'$ from the original feature set $\mathcal{S}$\;
% $k=1$ \;
%     \Repeat{$k$=K}{%
%     %   For each subset $S'$ calculate the value of $\frac{\sum_{U\in S'}SU(U,\boldsymbol{p}_{k})}{\sqrt{\sum_{U \in S'}\sum_{\mathcal{V} \in S'}SU(U,\mathcal{V}) }}$\;
%       Find the $\mathcal{S}^k$ through the Equation (\ref{featuresele.equ}) \;
%       $k=k+1$\;
%     }
%   }
%   \caption{Entropy-Based Feature Selection.}
%   \label{alg:fs}
% \end{algorithm}

% \look{The overall impression of (9)-(12) is an experiment report not a part of paper. Readers don't understand why to compute them. Why we need entropy? Why we need MI? Why we need symmetric uncertainty? How to interpret the final results in (12). We need to introduce the idea before computing entropy in (9).}

\subsection{The Relationship Between Load Patterns and Socioeconomic Factors}\label{subsec:prediction}

%Deep learning has achieved great performance in many areas \citep{cirecsan2012multi, wang2016does,li2015dnn}. 
% like image recognition, playing Go game, and automatic transition \citep{cirecsan2012multi, wang2016does,li2015dnn}. As defined in \citep{yosinski2014transferable}, deep learning is a technique that consists of computation models with multiple layers to learn the hidden information behind the data referring to the abstraction's multiple levels. 
% Comparing to the artificial neural network (ANN), the DNN model employs more complicated architectures and training process. Instead of having three layers, i.e. an input layer, a hidden layer, and an output layer as ANN, DNN can have numbers of layers with different numbers of neurons and parameters.    
Deep learning has been successfully used many energy studies, and
% \citep{ryu2017deep,hossen2017short,wang2017deep,wang2019novel}
the authors in \citep{zhang2018review} reviewed the research on deep learning applications to demand and renewable generation prediction, DR, and anomaly detection.
% \look{give more good references related to power and energy. Their authors may be the reviewers. I've removed \citep{cirecsan2012multi, wang2016does,li2015dnn} that are not relevant.} 
Different from these studies, we develop a DNN-based analytical model to uncover the relationship between the load patterns and socioeconomic factors. Thanks to the Pecan Street project \cite{street2016pecan}, both {household-level load data and socioeconomic data are made available and thus enable our work.} %Specifically, assuming that the utility company can collect the basic socioeconomic factors, e.g. ages, income, education levels, and so on, from the end-users, consumers' probability distribution of load patterns can be estimated by their socioeconomics through the model. 

\begin{figure}[t]
\centering
\includegraphics[width=1.0\columnwidth]{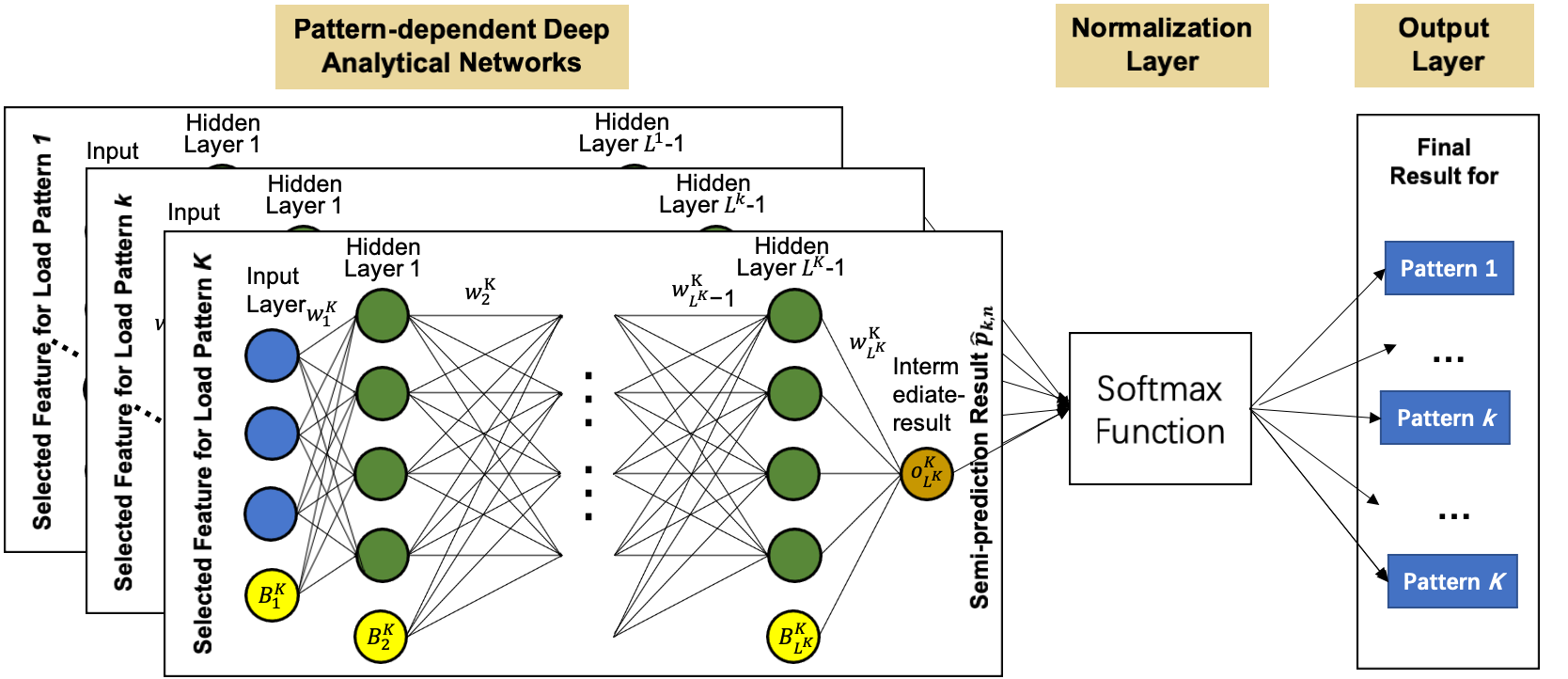}
\caption{The structure of the joint DNN model for revealing the relationship between load patterns and socioeconomic factors.}
\label{dnn} \end{figure}

From Section \ref{feature_sec}, we obtained the selected features $\mathcal{S}^k$ for each load pattern $k$. Accordingly, we build $K$ different DNN models with $\mathcal{S}^k$ as the input to estimate the probability distribution of load pattern $k$. However, in this way, the property of the probability distribution can not be always satisfied, and we introduce a normalization layer.
% the prediction of the probability distribution can not be always in the range of $[0,1]$, while the summation of the prediction 
%  of load patterns $p_{k,n}$ 
% of a certain consumer $n$ cannot be ensured to be consistently one. 
% to satisfy the property of the probability distribution.
% , which all conflict the natural law of the probability distribution. 
% We hence propose a prediction model separating the function of individually predicting the $k$-th load pattern distribution based on its selected features, jointly normalizing among the $K$ patterns to make the certain of the characteristic of probability distribution, and finally obtaining the prediction results, namely separate-to-joint structure. 
The structure of the model, represented in Fig. \ref{dnn}, is constructed with three components, e.g., pattern-dependent deep analytical networks, normalization layer, and output layer.
% \look{The order is wrong. This part and motivated structure design should be put in somewhere else. Otherwise, who knows what's the concern of the summation of different probability. We should write reasonably from the readers' perspective.} 
% We thus propose a novel separate-to-joint \look{is `separate-to-joint' a known term? If not, we cannot make it without proper explanation.} structure of the DNN model as represented in Fig. \ref{dnn}. Mainly constructing with three components, e.g. $K$ pattern-dependent deep prediction networks, normalization layer, and output layer, our proposed DNN model separates the function of individually predicting the $k$-th load pattern distribution based on its selected features, joint normalization among the $K$ patterns, and finally obtaining the prediction results.}
 
%  \look{The key problem is actually due to that we use different features for different patterns and thus need to build different DNNs. Then this raises the concerns of the summation of probability. We need to introduce the general idea at the beginning of the paragraph.}

% \rewrite{We employ a DNN model for each load pattern $k$ in the prediction networks. Each DNN model has the  pattern-corresponding feature set $\mathcal{S}^k$ as its inputs. The structure including weights and biases of each DNN model is identical. Each DNN model maps the features to the distribution for a certain load pattern. Shown at the bottom of Figure \ref{dnn}, the \textit{k}-th DNN model is taken as an example. In the model, we aim to map the value of features in $S^k$ belonging to consumer $n$ onto the pattern distribution value $p_{k,n}$. In particular, for each layer  $l\in\{1,\dots,L^k\}$, 
% \look{should this be `$=$' or `$\in$'?}
% an activation function is typically embedded to map the output from previous layer $o_{({L^k-1})}^k$ to the scalar state, which is also named as the output $o_{{L}^k}^k$ of layer $L^k$. The output of the previous hidden layer $o_{({L^k-1})}^k$ is thus treated as the input of the layer $L^k$. Combining with the inputs, the weight matrix $w_l^k$, and the bias $B_l^k$, the output of layer $o_l^k$ is computed by}
We employ a DNN model for each load pattern $k$ in the networks with the pattern-corresponding feature set $\mathcal{S}^k$ as the input, namely pattern-dependent deep analytical networks as shown in Figure \ref{dnn}. The structure including weights and biases of each DNN model is identical.
% Each DNN model maps the features to the distribution for a certain load pattern. 
We take the \textit{k}-th DNN model as an example. In the model, we aim to map the value of features in $S^k$ belonging to consumer $n$ onto the pattern distribution value $p_{k,n}$. {Concretely, for each layer $l\in\{1,\dots,L^k\}$, the outputs in the $(l-1)$-th layer are computed with the weights $\boldsymbol{w}_l^k$ through the inner product operation and then passed through a pre-defined activation function with bias $\boldsymbol{B}_l^k$. The calculation, 
% as shown in
\begin{equation}
     \boldsymbol{o}_{l}^k=\sigma \left(
    (\boldsymbol{w}_l^k)^\top \boldsymbol{o}_{l-1}^k+\boldsymbol{B}_l^k
    \right),
\end{equation}
generates the scalar state $\boldsymbol{o}_{l}^k$, which is also named as the output of layer $l$. 
% The output of the previous hidden layer $\boldsymbol{o}_{({l-1})}^k$ is treated as the input of the layer $l$.
We use the sigmoid function for the activation function $\sigma(\cdot)$.} We name the output of the last layer $\boldsymbol{o}_{L^k}^k$ the intermediate-result of the analysis, which is defined as $\hat{p}_{k,n}$. As each $\hat{p}_{k,n}$ is obtained distinctly, the intermediate-results in vector $\boldsymbol{\hat{p}}_{n}=[\hat{p}_{1,n},\dots,\hat{p}_{K,n}]$ 
% can not properly ensure the property of probability distribution value, e.g., the summation is consistently 1 and the value is in the range of [0,1].
can not be always in the range of $[0,1]$, while the summation of the prediction 
%  of load patterns $p_{k,n}$ 
of a certain consumer $n$ cannot be ensured to be consistently 1. 

% The normalization layer is thus utilized to satisfy the property of the probability distribution.
% \look{For all vectors and matrices, we need use bold fonts for the notations, e.g., $\boldsymbol{w}$ not $w$. Read some machine learning papers and use the standard formats. We cannot create something not commonly used by ourselves.}

% Additionally, in the light of the summation of the probability distribution covering all the \textit{k} pattern is one for each consumer, we introduce the normalization layer after the intermediate-result generated. Softmax function, with the ability to turn the input values into a vector whose value follows the probability distribution and sums up to 1, is greatly popular as a normalization algorithm \citep{bouchard2007efficient}.} It is therefore implemented as a joint activation function to the final prediction $\hat{p}_{k,n}^{\star}$. Along this line of consideration, the intermediate-prediction results $\hat{p}_{k,n}$ generated by the sub-prediction layer are aggregated and pooled as the input vector,
% \begin{gather}
%  I_{softmax}=
%   \begin{bmatrix}
%   \hat{p}_{1,n} \\
%   \dots\\
%   \hat{p}_{k,n} \\
%   \dots\\
%   \hat{p}_{K,n}
%   \end{bmatrix},
% \end{gather}  
% which is then passed to the fully connected softmax layer. Through the equation
%Additionally, in the light of the probability distribution is positive, the summation of which covering all the \textit{k} pattern is one for each consumer, % We thus introduce the normalization layer 

The normalization layer is thus used to enforce the property of the probability distribution 
after the intermediate-result $\boldsymbol{\hat{p}}_{n}$ is generated. {The softmax function \citep{bouchard2007efficient}, as a popular normalization algorithm, is employed to transform the input vector $\boldsymbol{\hat{p}}_{n}$ into $\boldsymbol{\hat{p}}_{n}^{\star}=[\hat{p}_{1,n}^{\star},\dots,\hat{p}_{K,n}^{\star}]$ as
\begin{equation}\label{eq:softmax}
    \hat{p}_{k,n}^{\star}=\frac{e^{\hat{p}_{k,n}}}{\sum_{k=1}^{K}e^{\hat{p}_{k,n}}},
\end{equation}
whose value satisfies the property of probability distribution.}
% \look{Are we using (14) in the model? If so, we have a big problem.}
% the softmax function computes the probability distribution over the input labels to emphasize the higher value while weakening the lower value. The output $\hat{p}_{k,n}^{\star}$ is then defined as the prediction result of the user \textit{n} in \textit{k}-th load pattern. Therefore, during the training stage, $\hat{p}_{k,n}^{\star}$ is involved in the loss function to tune the parameters of the \textit{k}-th pattern-dependent deep network.
The output $\hat{p}_{k,n}^{\star}$ is defined as the evaluated percentage of user \textit{n}'s load profiles in \textit{k}-th load pattern. The normalization layer guarantees that the evaluated probabilities $\hat{p}_{k,n}^{\star}$ are non-negative and within $[0,1]$, and their summation is $1$. During the training stage, $\hat{p}_{k,n}^{\star}$ is used to define the loss function for the \textit{k}-th pattern-dependent deep network.

We use the mean squared error (MSE) as the loss function to train $K$ DNN models in parallel. By measuring the dissimilarity between the original probability $p_{k,n}$ and $\hat{p}_{k,n}^{\star}$, the loss function is calculated by
\begin{equation}
    MSE_k=\sqrt{\frac{1}{N'}\sum_{n=1}^{N'} (p_{k,n}-{\hat{p}_{k,n}^{\star}})^2},
    \label{MSEeq}
\end{equation}
where $N'$ is the number of consumers randomly selected to train the model from the \textit{N} consumers. Stochastic gradient descent method is introduced to train and update the weight $\boldsymbol{w}_l^k$ and bias $\boldsymbol{B}_l^k$  in the back-propagation way. And the training stops when $MSE$ is below some threshold or the number of iterations or epochs is above some threshold. %backpropagatively.

To show the improvement made by our proposed analytical scheme, we compare our model with a regression model and two DNN-based models as benchmarks:
\begin{itemize}
 \item \textit{Benchmark 1} employs XGBoost to construct a non-linear regression model. The model is to describe the relationship between the selected features with non-selected features in $\mathcal{S}$ as the indicator.
    \item \textit{Benchmark 2} uses a single DNN model to using non-selected features in $\mathcal{S}$ as inputs.
    \item \textit{Benchmark 3} has the same structure of our analytical model, as shown in Fig. \ref{dnn}, but takes non-selected features in $\mathcal{S}$ as inputs.
   
\end{itemize}
The comparison results will be present in the following section.

% \textcolor{violet}{Summarily, we construct our prediction model through $K$ pattern-dependent deep prediction networks individually. We also count in the natural property of the predicted probability distribution jointly in the model by involving the Softmax normalization. The back-propagating tuning subsequently makes effects on the respective networks through the loss function.}
% \look{Please rewrite this paragraph.}

\section{Simulation Results and Discussions}
\label{sec:simulations}
% \look{We should add several sentences here to describe what we are going to present for Section IV before directly going to Subsection A.}
% \look{I also stop here and leave Section IV to you to revise. I will follow up on Friday.}

This section presents the load patterns clustered by the K-Medoids algorithm for weekdays and weekends, respectively. Then we select features using the entropy-based algorithm and employ Pearson correlation to further interpret the result. Last but not least, the overall performance of our proposed analytical model is validated through comparisons with three benchmarks. Before presenting the numerical results, we describe the data used in this work. {We use the smart meter data in 2019 and socioeconomic information from Pecan Street \cite{street2016pecan}.} After pre-processing the data, we selected $433$ households with the complete hourly consumption for three years (2015-2017) and the corresponding socioeconomic information.
% \vspace{-0.5cm}
% the proposed analysis procedures on consumer load patterns and deep learning models on the pattern and socioeconomic information are evaluated. The correlation between load patterns and socioeconomic information is studied and present in detail with the realistic consumption data. The performance of the probability distribution of load pattern prediction is well represented.   }
%\vspace{-0.5cm}

% \subsection{Data Description}
% To instantiate the evaluation framework, we primarily leveraged the residential consumption data and economic information from the Pecan Street database \cite{street2016pecan}. 
% The database includes load data for more than 1000 households with the resolutions from 1-hour down to 15-second, where the per-appliance power demands are counted as well. Moreover, the house-related socioeconomic information, e.g. age of resident, annual income, educational level, and the total square footage of the household, are provided. Of these, we further selected 433 households with the complete hourly total consumption data from the year 2015-2017 and their socioeconomic information. 

\subsection{Load Pattern by Clustering}
\begin{figure}[!tbp] 
\centering
\includegraphics[width=1\columnwidth]{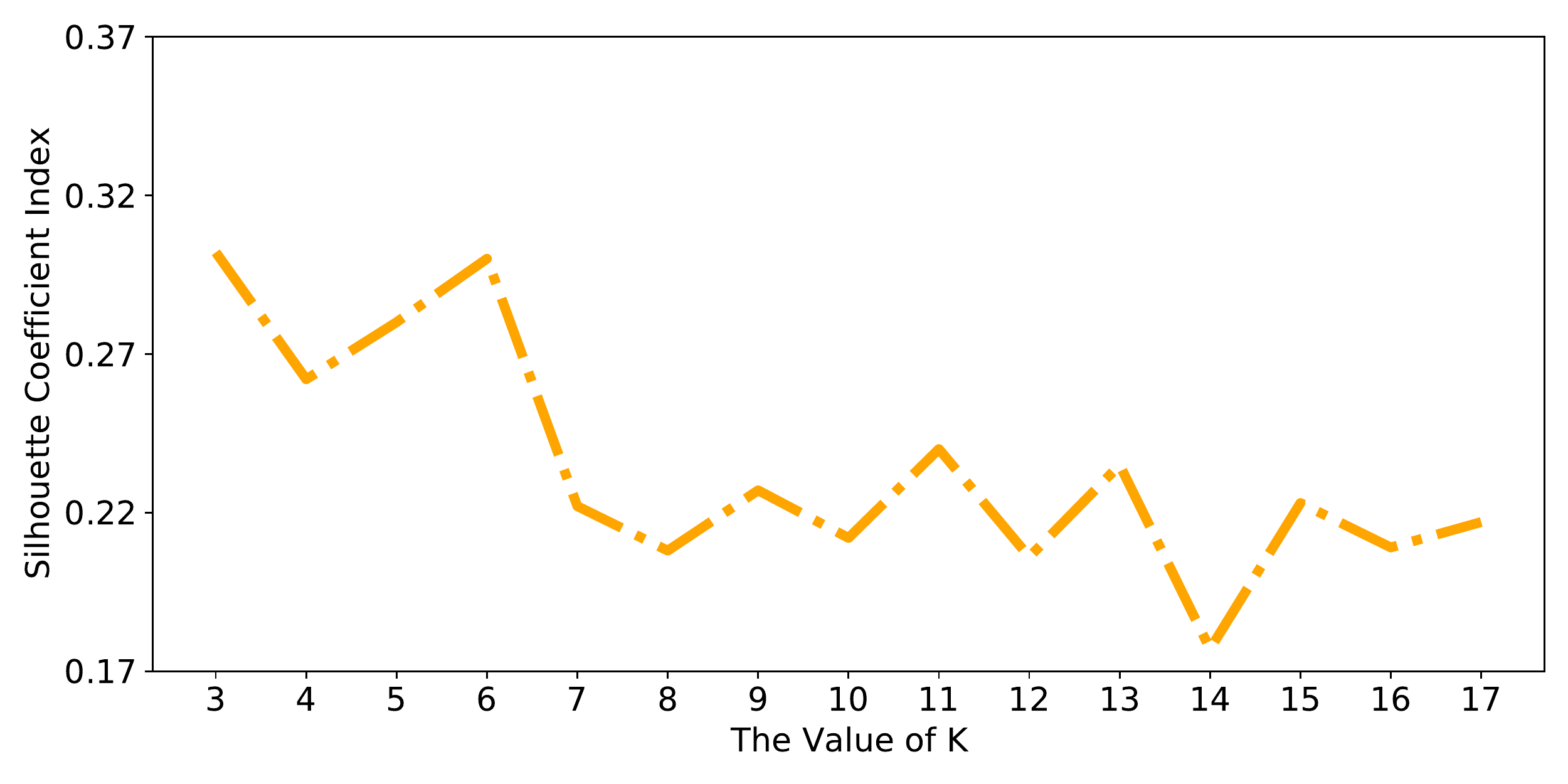}
\vspace{-0.3cm}
\caption{The $SC$ index according to the change of $K$.}
\vspace{-0.5cm} 
\label{SC} \end{figure}

% The consumption data are divided into two parts, i.e., `weekday' and `weekend' labeled by $e$ and $w$. We normalize the original load profiles as $L_{w,d,t}^n$ and $L_{e,d,t}^n$ for weekdays and weekends, respectively. 
% and then determine the best number of $K$. 
{
We evaluate the clustering performance of the K-Medoids 
method under different values of $K$ and calculate the Silhouette Coefficient by Equation \eqref{SCeq}. The results 
% illustrating Silhouette Coefficient value varying according to $K$ obtained by K-Means and K-Medoids method separately 
are represented as the average values of `weekday' and `weekend' load clustering in Fig. \ref{SC}.
% , which are also categorized by . 
% Compared with K-Means, the $SC$ value calculated by K-Medoids performs a more distinct result. Neverthless, the number of clusters $K=6$ using Silhouette Coefficient Index in Equation (\ref{SCeq}) \ref{SC}. 
% We determined $K=6$ for the clustered load patterns. 
%  The values of the index are calculated by varying the number of $K$ in the clustering algorithm, namely, each run of the clustering algorithm generates a $SC$ index to represent the appropriate level measurement. 
% We use developed K-Medoids method in Algorithm \ref{alg:kmedoid} to cluster normalized load profiles. 
% We have the $SC$ value when K-Medoids applied to weekday/weekend data rendered in the figure. Moreover, 
We see that the K-Medoids method has distinct values of the Silhouette Coefficient with respect to $K$.
% and achieves the highest score compared with K-Means on the same dataset. 
The clustering result with $K=6$ performs the best with the highest average value
% while using K-Medoids 
for both weekday and weekend. Meanwhile, $K=3$ also demonstrates a high value. However, considering the interpretability of the clustering results, we choose $K=6$.}
% For the weekend data, the top 3 values of $SC$ are obtained by K-Medoids when $K=$3, 5, and 6 individually. 
In the following discussion, we represent $6$ load patterns on weekdays and weekends and distinguish G1-G6 for weekdays and weekends by `W’ and `E’, respectively.

% The performance of the clustering under diverse $K$ value is evaluated in Fig. \ref{SC}. We have the $SC$ value when K-Medoids applied to weekday/weekend data rendered in the figure. Moreover, to have the comparison, the $SC$ obtained by the K-Means is evaluated in the same figure as well. We can tell from the figure that K-Medoids method can have more distinct results within the change of $K$ and achieve the highest score compared to the K-Means method applying to the same dataset. 
% The clustering with $K=6$ performs the best while applying K-Medoids method to the weekday category. For the weekend data, the top 3 values of $SC$ are obtained by K-Medoids when $K=$3, 5, and 6 individually. In the following discussion, we represent 6 load patterns both on weekdays and weekends and distinguish G1-G6 of two categories by using `W’ and `E’.

% \begin{figure}[tp!] 
% \centering
% \includegraphics[width=0.9\columnwidth]{Silhouette_Coefficient_Index.pdf}
% \vspace{-0.5cm}
% \caption{The silhouette coefficient index according to the change of $K$.} 
% \vspace{-0.5cm}
% \label{SCfig}
% \end{figure}
 
%  Basic observations on the encoded load shapes from the population data are described. Fig. 11 shows the most frequent 16
% load shapes which account for more than 1% of load shapes in
% whole encoded data with the final dictionary. It can be seen that
% most of high usage happens in the late afternoon or evening,
% which represents the lifestyle of usual households. A sample
% lifestyle may be that households leave home in the morning,
% come back home after school or work and consume electricity
% till they sleep. Notice that many shapes can be differentiated
% by the timing of peak consumption, indicating this might be a
% good variable to design programs around. We can compare the
% number of households that have the top 16 load shapes among
% their top 5 load patterns to confirm that the top 16 patterns represent a population and not just a small set of consumers. Fig. 12
% shows that on average each top 16 load shape appears in the top
% 5 of 15.3% of households.

\begin{figure*}[!t] 
\centering
\includegraphics[width=0.8\textwidth]{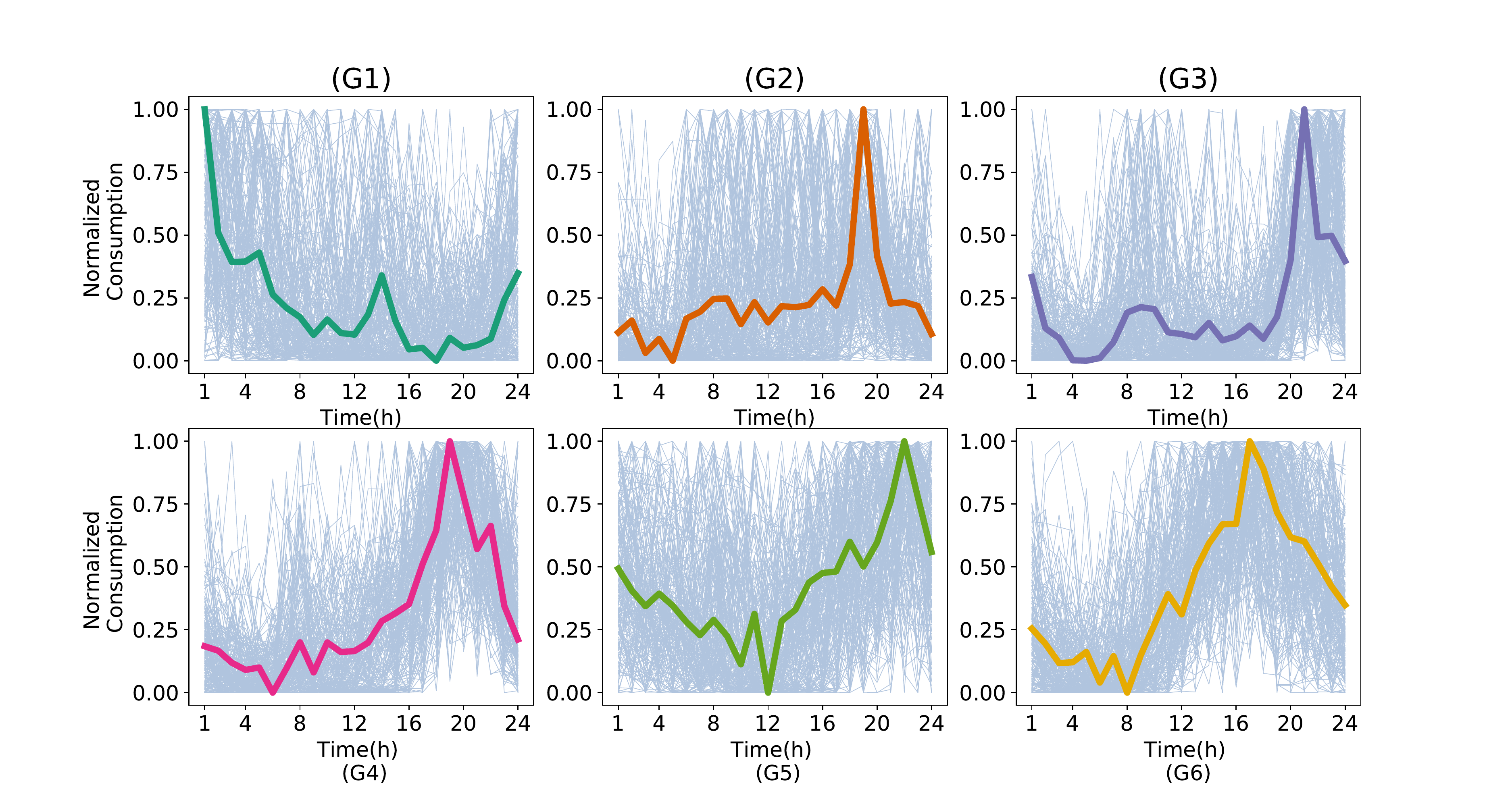}
%\vspace{-0.5cm}
\caption{The six load patterns for weekdays.}
\label{pattern1} 
%\vspace{-0.3cm}
\end{figure*}

\begin{figure*}[!t] 
\centering
\includegraphics[width=0.8\textwidth]{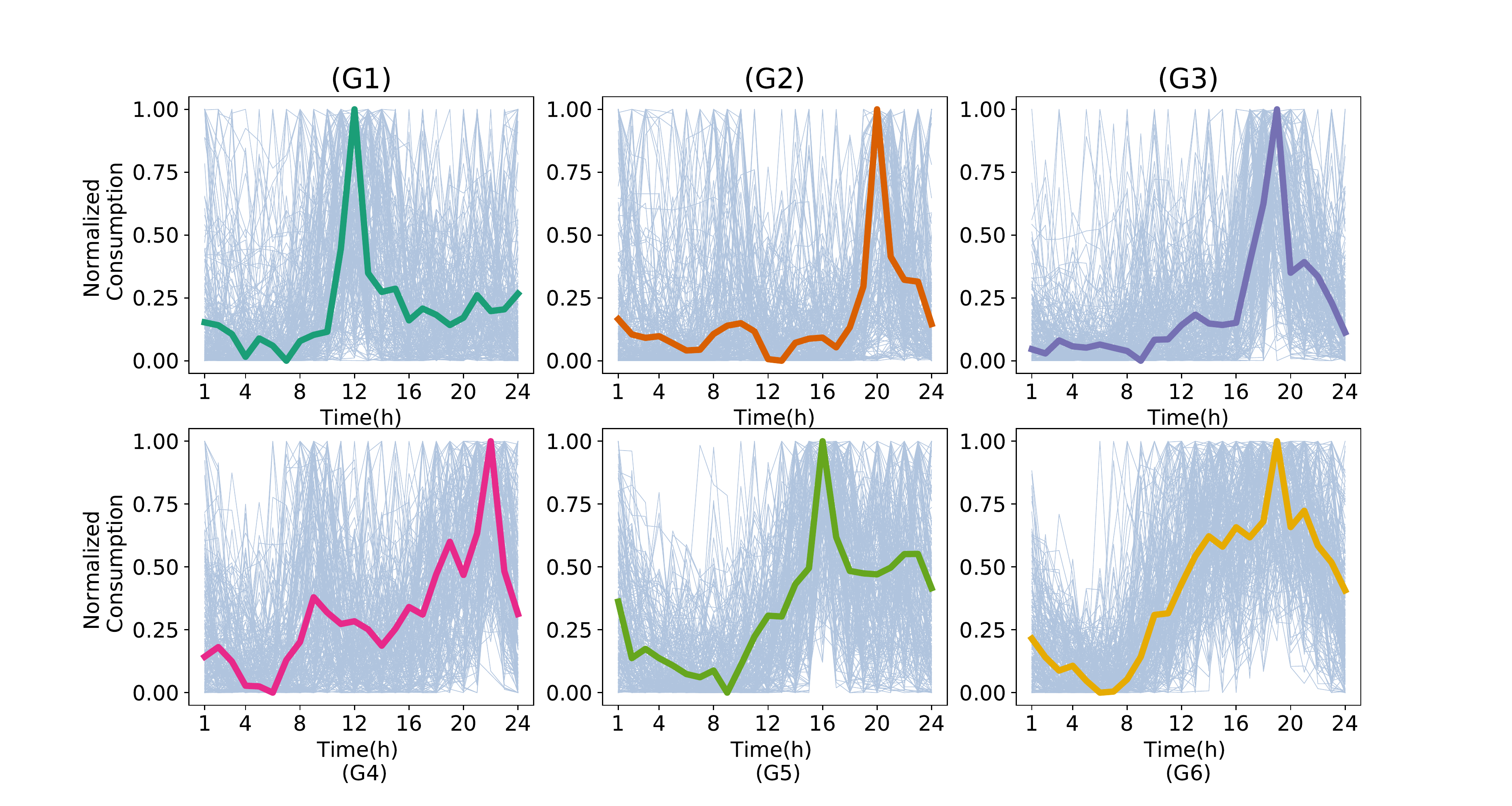}
%\vspace{-0.9cm}
\caption{The six load patterns for weekends.}\label{pattern2} 
%\vspace{-0.5cm}
\end{figure*}

We show the clustered load patterns in Fig. \ref{pattern1} for weekdays and Fig. \ref{pattern2} for weekends, where the blue dash lines display the medoids and the grey lines depict all load profiles in the corresponding cluster.
% At first glance on the medoids,
The main difference between weekday and weekend load patterns lies in the load shape, especially in peak time. We see that all $6$ medoids on weekdays have evening peaks.
% , though some load profiles have peaks in the daytime.}
On the contrary, we find that the shapes of load patterns on the weekend are more diverse, e.g., peaks in both daytime and evening, and relatively higher consumption during the daytime, indicating less routine household activities on weekends.

% such a late peak does not appear in every pattern of the weekend.  The weekend patterns lie clearer diving lines between peak and off-peak consumption, e.g. greater variation among each time interval.}

% \begin{figure}[htbp]
% \centering
% \begin{minipage}[t]{0.48\textwidth}
% \centering
% \includegraphics[width=1\columnwidth]{work_group.pdf}
% \vspace{-0.7cm}
% \caption{The six load patterns of weekday.}
% \vspace{-0.5cm}\label{pattern1} 
% % \end{figure}
% \end{minipage}
% \begin{minipage}[t]{0.48\textwidth}
% \centering
% \includegraphics[width=1\columnwidth]{holi_group.pdf}
% \vspace{-0.7cm}
% \caption{The six load patterns of weekend.}\vspace{-0.5cm} \label{pattern2}
% \end{minipage}
% \end{figure}

% \begin{figure}[htbp]
%     \centering
%   \vspace{-0.7cm}
%     \subfigure[weekday\vspace{-0.7cm}]{
%         \includegraphics[width=1\columnwidth]{work_group.pdf}
%         \label{pattern1} 
%     }
    
%     \quad    %用 \quad 来换行
%     \vspace{-0.7cm}
%     \subfigure[Weekend\vspace{-0.7cm}]{
%     	\includegraphics[width=1\columnwidth]{holi_group.pdf}
%         \label{pattern2} 
%     }
%     \vspace{-0.7cm}
%     \caption{This is a Demo of $2\times 2$}
%     % \vspace{-0.7cm}
%     % \label{fig.1}
% \end{figure}

\begin{figure}[tbp!] 
\centering
\includegraphics[width=0.95\columnwidth]{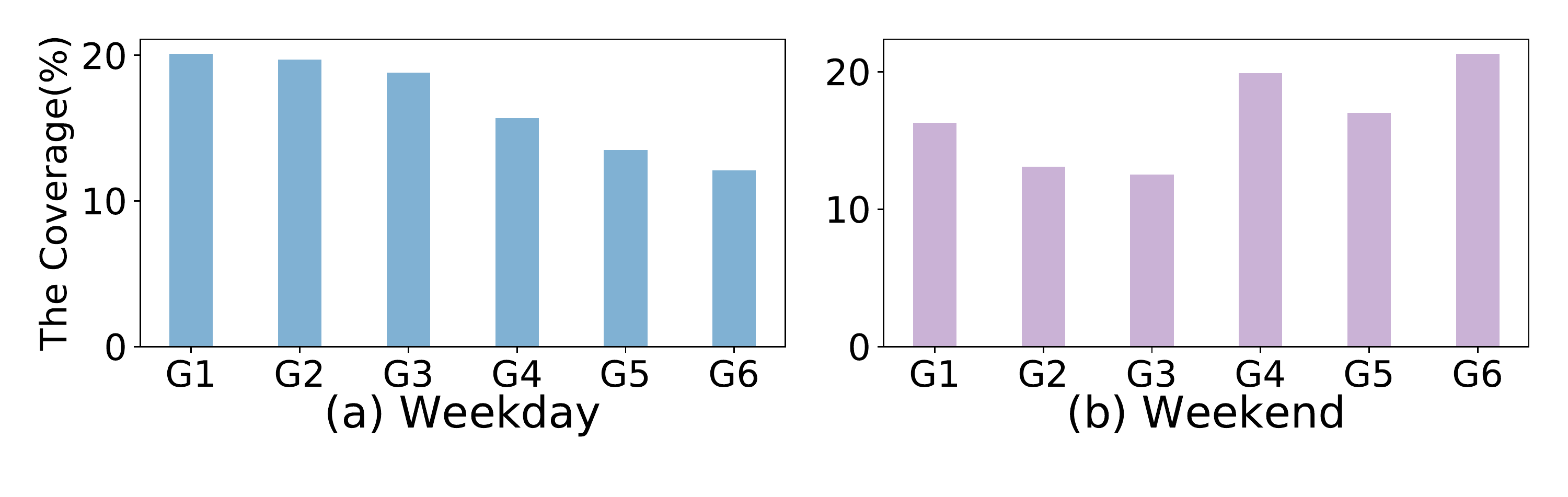}
%\vspace{-0.5cm}
\caption{The percentage distribution for load patterns on Weekdays and Weekends.}
\label{piechart} 
%\vspace{-0.5cm}
\end{figure}

Meanwhile, we show the percentage of consumers’ daily load profiles belonging to each load pattern in Fig. \ref{piechart}. For the load patterns on weekdays, we see that G1, G2, and G3 are the most representative load patterns covering 60\% of load profiles. This indicates a lifestyle of consumers in general that residents leave home in the morning, come back from work or school in the evening, and the energy consumption goes to a daily peak until sleeping. 
% Or, we can also interpret these household may use solar energy primarily when the sun is up.
Note that the main difference among G1, G2, and G3 is the peak time.
%  From the shapes of the profiles, we can conjecture that
% the resident of these households may be the persons who
% work regularly during daytime. 
% Compared with G2 and G3, G1 has a more flat shape, especially during the daytime. Different from G1, G2 and G3 have a quite similar shape but different peak times, where the peak is happened during afternoon or late  afternoon for G2 and during evening for G3 separately. G3, exhibiting the percentage of 15\%, has similar peak and off-peak time as G2, but its semi-peak in late evening shows the difference. 
As two relatively rare consumption types, G5 and G6 cover 13.5\% and 12.1\% of weekday load profiles. Compared with G1-G3, G5 and G6 have higher consumption during the daytime, implying regular occupancy in the daytime on weekdays. The distribution of load patterns on weekend can also be found in Fig. \ref{piechart}(b). We see that G4, G5, and G6 are the most representative load patterns on weekends. 
% With less greatly variant energy-consuming from daytime till the night, the shapes of the three load patterns infer that people at the weekend are more preferable to stay at home.
Compared with the top three representative load patterns on weekdays, those on weekends have a single peak or dual peak during 9:00-18:00.
% Comparing with the familiar lifestyle of the weekday, people at the weekend seems more preferable to stay at home with less variant energy consuming.
% We can tell the similarity between the G6 in weekday and G6 in weekend besides the peak of G6 in weekend is a bit later. However, the frequency of the weekend G6 appearance is visually much higher. We detect other similar 'twins' are well judged not by the representative coverage but in the basis of the shape, e.g. G3 in weekday and G5 in weekend, G2 in weekday and G1 in weekend. Likewise, the peak shifts to the daytime in G1 and G5 load patterns in weekend in the comparison with their twins in weekday category. 
% Conclusively, not only from the load shapes but also the coverage, the representative load patterns captured for the weekday and weekend are significantly different. 
% \textcolor{}{Specifically, the weekday patterns mostly reveal the out-to-work lifestyle
% % figured by the consumption behavior where most of the peak happened during the late of the day,
% where the weekend load patterns estimate more kinds of lifestyles.}% makes the effect on weekend patterns resulting in even more different shapes of patterns.}

\subsection{Pattern-Related Feature Selection}\label{simfeacture}
In this subsection, we show the selected socioeconomic features for each load pattern.
% (6 for the weekday and 6 for the weekend)
All the features in the original set $\mathcal{S}$ are listed in the first column of TABLE \ref{feature_sele}. Applying the feature selection method presented in Section \ref{feature_sec}, we obtain the selected features in $\mathcal{S}^k$ that are marked.
% is selected from $\mathcal{S}$ which  The features 
% is selected by the entropy-based feature selection methodology and the features in the subset are
% in TABLE \ref{feature_sele}.
% where the age higher than 65 in particular during weekday and the younger people, e.g. age range 13-24 and 25-49, during weekend.

\newcommand{\tabincell}[2]{\begin{tabular}{@{}#1@{}}#2\end{tabular}}

\begin{table}[!tp]
\raggedleft
% \vspace{-0.2cm}
\caption{The Selected Feature for Each Load Pattern}
% \vspace{-0.3cm}
\label{feature_sele}
\resizebox{\columnwidth}{18mm}{
\begin{tabular}{|c|c|c|c|c|c|c|c|c|c|c|c|c|c|}
\hline
\multicolumn{2}{|c|}{} & \multicolumn{6}{c|}{\textbf{Weekday}} & \multicolumn{6}{c|}{\textbf{Weekend}} \\ \cline{3-14} 
\multicolumn{2}{|c|}{\multirow{-2}{*}{\textbf{Feature}}} & G1 & G2 & G3 & G4 & G5 & G6 & G1 & G2 & G3 & G4 & G5 & G6 \\ \hline
 & under 12 &  & \cellcolor[RGB]{244,208,0} & \cellcolor[RGB]{244,208,0} &  &  &  &  &  &  &\cellcolor[RGB]{244,208,0}  &  &  \\ \cline{2-14} 
 & 13-24 &  &  &  & \cellcolor[RGB]{244,208,0} &  &  &  & \cellcolor[RGB]{244,208,0} & \cellcolor[RGB]{244,208,0} &  &  & \cellcolor[RGB]{244,208,0} \\ \cline{2-14} 
 & 25-49 &  &  &  & \cellcolor[RGB]{244,208,0} &  &  &  &\cellcolor[RGB]{244,208,0}  &  &  & \cellcolor[RGB]{244,208,0} &  \\ \cline{2-14} 
 & 50-64 & &  &  &  & \cellcolor[RGB]{244,208,0} & \cellcolor[RGB]{244,208,0} &  &  &  &  & \cellcolor[RGB]{244,208,0} &  \\ \cline{2-14} 
\multirow{-5}{*}{\rotatebox{90}{\tabincell{c}{Number\\ of Resident \\ Age Range}}} & \shortstack{over 65} & \cellcolor[RGB]{244,208,0} & \cellcolor[RGB]{244,208,0} & \cellcolor[RGB]{244,208,0} &  &  & \cellcolor[RGB]{244,208,0} &  &  &  & \cellcolor[RGB]{244,208,0} &  &  \\ \hline
\multicolumn{2}{|c|}{\shortstack{Education\\ Level}} & \cellcolor[RGB]{244,208,0} & \cellcolor[RGB]{244,208,0} &  & \cellcolor[RGB]{244,208,0} &  &  & \cellcolor[RGB]{244,208,0} &  & \cellcolor[RGB]{244,208,0} & \cellcolor[RGB]{244,208,0} &  & \cellcolor[RGB]{244,208,0} \\ \hline
\multicolumn{2}{|c|}{\shortstack{Annual \\Income}} &  &  &  &  &  &  &  &  &  &  & \cellcolor[RGB]{244,208,0} &  \\ \hline
\multicolumn{2}{|c|}{\shortstack{Total Square \\ Footage}} &  &  &  &  &  &  & &  &  &  &  &  \\ \hline
\end{tabular}}
%\vspace{-0.5cm}
\end{table}

% It can be validated that the feature ‘Total Square Footage’ does not show a strong correlation with consumers’ load patterns which can be explained that the load profiles have been normalized.
% , as we focus on the temporal variation of consumption. 
% In contrast, 
We see that for all the load patterns, no matter on weekdays or weekends, the most related feature(s), i.e., the ones in $\mathcal{S}^k$, are very different. Taking G1(W) and G2(W) as an example, two features, e.g., \textit{Age Over 65} and \textit{Education Level}, are selected simultaneously for $\mathcal{S}^{G1(W)}$ and $\mathcal{S}^{G2(W)}$, but the feature \textit{age under 12} is selected in $\mathcal{S}^{G2(W)}$ alone. The diverse pattern-related feature sets reveal the importance of the feature-dependant design for the prediction model. In addition, age and education are highly influencing features, suggesting that age and education help determine the load patterns.
% play a key role in consumption behaviors. % It can be observed from G1, G2, and G3, the top 3 representative load patterns in the weekday, that consumers with the age older than 65 makes the great impact.
Specifically, we find that the age greater than $65$ exhibits a high correlation with load patterns on weekdays.
% and are potentially the target participants in demand response programs. 
Meanwhile, on weekends, the age ranging from $12$ to $49$ has a higher impact on the load pattern.
% It can somehow explain the diverse peak time on weekends that young peoples have more activities.
Besides, the education level shows a strong impact on all the load patterns. In contrast, the feature \textit{Total Square Footage} does not show a strong correlation with any load patterns, because the load profiles have been normalized in this study. The selected features for load patterns on weekdays and weekends are different, which justifies the need to consider weekday and weekend load patterns separately.

\subsection{The Correlation Coefficient between The Features} 
\begin{figure}[tbp!] 
\centering\includegraphics[width=1\columnwidth]{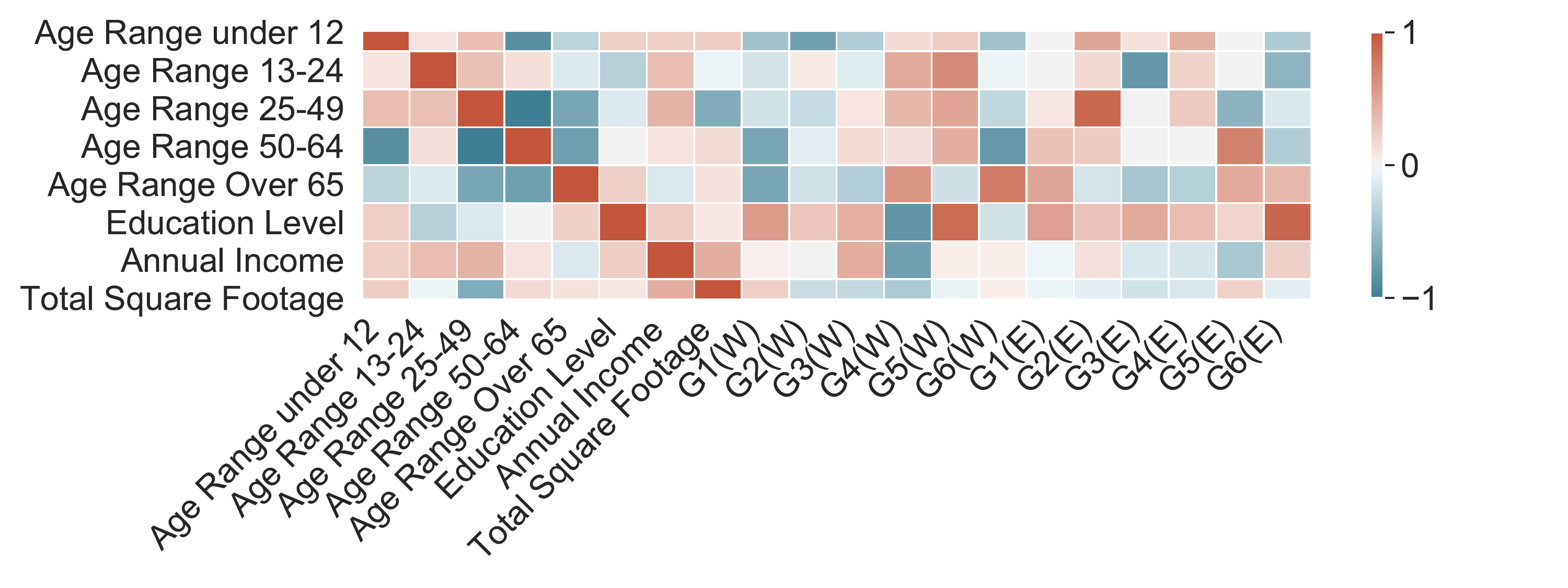}
%\vspace{-1cm}
\caption{The Pearson correlation coefficient between the features and load patterns.}
\label{hm} 
%\vspace{-0.5cm}
\end{figure}

We further 
% analyze the pairwise correlation 
% between features 
use Pearson correlation coefficient
% \begin{equation}
%     \rho_{U\in S,V\in S}=\frac{\EX((U-\mu_U)(V-\mu_V))}{\tau_U \tau_V}
%     \label{pearson}
% \end{equation}
to provide more insights into the pairwise correlation not only between the features and load patterns but also among different features.
% Even though the Pearson correlation is more feasible for the data following the normal distribution, we employ it to simply represent the correlation and support our feature selection result. 
The coefficients represent the dependence of the relationship, while the sign exhibits a positive or negative correlation. We create a heatmap to visualize the correlations in Fig. \ref{hm},
% visualize the value in Fig. \ref{hm} by 
in which we use the red gradient color for varying degrees of positive correlations and blue for negative correlations.
% As we figure the patterns from weekday and weekend in the same heatmap, the load patterns for weekdays are distinguished by adding 'W' after G1-G6 and for weekend adding 'E'.
Taking the feature \textit{Age Range Over 65} as an example, it has a negative correlation with \textit{Age Range 50-64} and \textit{Age Range 25-49}, which indicate that the residents from these three age ranges are unlikely to live in the same household. Given such a strong correlation, the feature \textit{Age Range Over 65} is thus selected as the other two features are redundant. The selected features for G1(W) are consistent with this analysis. Even though \textit{Age Range 50-64} shows a strong correlation with G1(W) in the heatmap, it is not selected in TABLE \ref{feature_sele}. The above results also validate the employed feature selection method that does count in the correlation measurement and apparent redundancy.
%\vspace{-0.5cm}

%   \begin{figure}[tph] 
% \raggedleft\includegraphics[width=13cm]{holi_heatmap.pdf}
% \caption{The structure of joint DNN model for load pattern prediction.} \label{fig2} \end{figure}

% \begin{figure}[htbp]
% \centering
% \subfigure[pic1.]{
% \includegraphics[width=13cm]{work_heatmap.pdf}
% %\caption{fig1}
% }
% \quad
% \subfigure[pic2.]{
% \includegraphics[width=13cm]{holi_heatmap.pdf}
% }
% \caption{ pics}
% \end{figure}
\subsection{Load Patterns and Socioeconomic Factors} 
We take the selected features in 
% Section \ref{simfeacture}
TABLE \ref{feature_sele} as the inputs of our analytical model presented in Section \ref{subsec:prediction} to reveal the corresponding probability distributions of load patterns.
% The prediction models are separated for the weekday and weekend. 
{For both weekday and weekend data, we use 70\% of the data for training, 15\% for validation, and the remaining 15\% for the test. By applying grid search \cite{bergstra2011algorithms} on the validation dataset and MSE as the evaluator, we optimize the hyper-parameters, e.g., set $5$ layers and $1024$ neurons for each layer.}

% To show the improvement made by feature selection and to validate our proposed prediction model, we compare our prediction model with two DNN-based models as benchmarks. In all the compared models, we use the same structure for the hidden layer. Specifically, for benchmark 1, instead of using our proposed $K$ pattern-dependent networks, we adopt one unified DNN model to predict all the probability distributions of load patterns at once using non-selected features in $\mathcal{S}$ as inputs. Benchmark 1 also serves as the baseline for the comparison. For benchmark 2, we use the same structure 
% of our prediction model as shown in Fig. \ref{dnn} but
% % (6 separate DNNs for 6 load patterns combined with a normalization layer) 
% take non-selected features in $\mathcal{S}$ as inputs.
% Comparing with the benchmark 2, we aim to estimate the performance improved by feature selection. }

\subsubsection{Individual Households}
\begin{figure*}[!th] 
\centering\includegraphics[width=1.2\columnwidth]{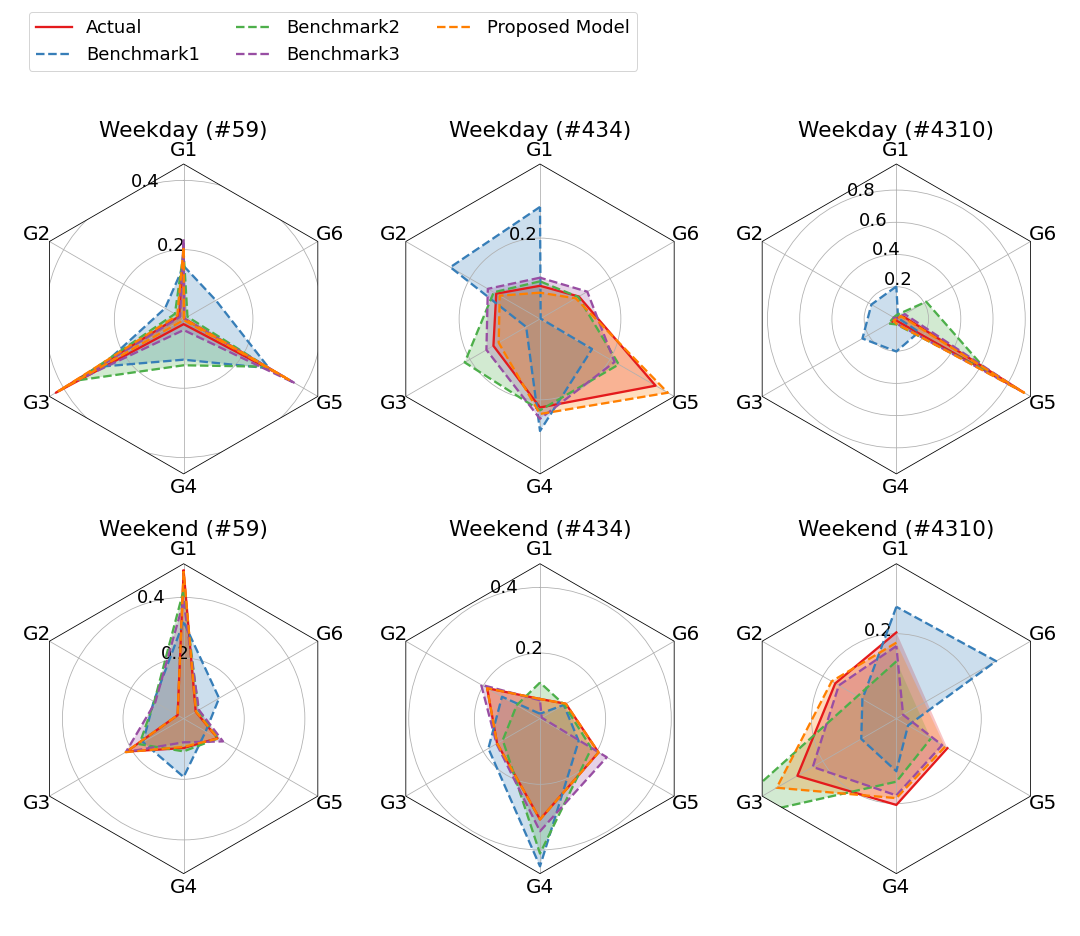}
\caption{Comparison of our model and three benchmarks for estimating load pattern distributions of three households (\#59, \#434, and \#4310).}
% \look{please also change this .png file to .pdf file.}
\label{radar} 
\end{figure*}

{We select three households (\#59, \#434, and \#4310) from the dataset \cite{street2016pecan} to show the test results, and the household information is listed in Table \ref{tb:infor}.}
% to have an intensive study. 
{The three households have diverse socioeconomic features, including two seniors with high education levels and income in household \#59, a family of two children and two adults with medium income in household \#434, and three adults with the college education and modest income in household \#4310. }
% cataloged in Table \ref{Exam_pre} as individual examples. From the socioeconomic information, we can tell these three are entirely different.
% with different numbers of residents aging under different levels, education level, annual income, and total square footage of the living space. 
% Therefore, it is believed the accuracy of prediction results 
% of distribution probability of load patterns
% can, by and large, image the whole situation.
% We depict the prediction results still using the 2 benchmark models and our proposed model.
% in the weekday and at the weekend, respectively.
The results are depicted in the radar maps of Fig. \ref{radar}, illustrating the estimated distributions of load patterns for our model and three benchmarks.
% The red dash line shows the ground truth of the percentages of each load pattern and the blue, green and purple represent the prediction results of benchmark 1, benchmark 2, and proposed model distinctly. 

We see in Fig.~\ref{radar} that the regression results (in orange dash-line)using our proposed method best match the ground truth (in red line) for all three case studies on weekdays and weekends. In contrast, Benchmark 1's results (in blue dash-line) deviate from the ground truth significantly in all the cases, suggesting that the regression model is underfitting. Benchmark 2's results (in green dash-line) and Benchmark 3's results (in purple dash-line) are better than that of Benchmark 1 but not competitive to our proposed method. Specifically, significant inaccuracy occurs in Benchmark 2 for \#59 and \#4310 on the weekend.

The results reveal rich information about how different households consume energy. For example, Household \#59 has two elderly residents with high education levels and high incomes. They have a stable consumption behavior with three major load patterns in G1, G3, and G5. On weekdays, they may cook lunch leading to a midday peak in G1(W), getting up early in the morning and cook dinner in G3(W), or stay at home with high consumption in the afternoon followed by a night peak in G5(W). During the weekend, the dominant pattern G1(E) indicates that they have a midday peak, suggesting energy-intensive household activities. They also have energy-intensive household activities in the afternoon (G5(E)) or for dinner (G3(E)). Similarly, the load patterns of Households \#434 and \#4310 can be explained.

% , as no feature selection is considered in this model
% Conclusively, it is demonstrated by the test results that combining with feature selection the deep learning prediction model effectively improves the prediction accuracy.}

% Please add the following required packages to your document preamble:
% \usepackage{multirow}
% \begin{table}[tp!]
% \centering
% \caption{The Socioeconomic features of three consumers.}
% \vspace{-0.2cm}
% \label{Exam_pre}
% \begin{tabular}{|c|c|c|c|}
% \hline
% \multirow{2}{*}{\begin{tabular}[c]{@{}c@{}}Socioeconomic \\ Features\end{tabular}} & \multicolumn{3}{c|}{Consumer ID} \\ \cline{2-4} 
%  & 59 & 434 & 4310 \\ \hline\hline
% Age Range under 12 & 0 & 2 & 0 \\ \hline
% Age Range 13-24 & 0 & 0 & 2 \\ \hline
% Age Range 25-49 & 0 & 2 & 0 \\ \hline
% Age Range 50-64 & 0 & 0 & 1 \\ \hline
% Age Range Over 65 & 2 & 0 & 0 \\ \hline
% \shortstack{Education \\ Level} & \shortstack{Postgraduate \\Degree} & \shortstack{Postgraduate \\Degree} & 3 \\ \hline
% \shortstack{Annual \\ Income }& \ \shortstack{\$300,000 - \\ \$1,000,000} & \shortstack{\$100,000 \\ - \$149,999} & 7 \\ \hline
% Total Squared Footage & 3830 & 2160 & 3130 \\ \hline \hline
% \end{tabular}
% Please add the following required packages to your document preamble:
% \usepackage{multirow}

\begin{table}[]

\caption{The Information of Three Selected Households}
\begin{tabular}{|c|c|c|c|c|}
\hline\hline
\multicolumn{2}{|c|}{\multirow{2}{*}{\begin{tabular}[c]{@{}c@{}}Socioeconomic\\ Information\end{tabular}}} & \multicolumn{3}{c|}{Consumer ID} \\ \cline{3-5} 
\multicolumn{2}{|c|}{} & \#59 & \#434 & \#4310 \\ \hline \hline
\multirow{5}{*}{\rotatebox{90}{Age Range}} & under 12 & 0 & 2 & 0 \\ \cline{2-5} 
 & 13-24 & 0 & 0 & 2 \\ \cline{2-5} 
 & 25-49 & 0 & 2 & 0 \\ \cline{2-5} 
 & 50-64 & 0 & 0 & 1 \\ \cline{2-5} 
 & over 65 & 2 & 0 & 0 \\ \hline
\multicolumn{2}{|c|}{\begin{tabular}[c]{@{}c@{}}Education\\  Level\end{tabular}} & \begin{tabular}[c]{@{}c@{}}Postgraduate\\ Degree\end{tabular} & \begin{tabular}[c]{@{}c@{}}Postgraduate\\ Degree\end{tabular} & \begin{tabular}[c]{@{}c@{}}College \\ Graduate\end{tabular} \\ \hline
\multicolumn{2}{|c|}{\begin{tabular}[c]{@{}c@{}}Annual  \\ Income (\$)\end{tabular}} & \begin{tabular}[c]{@{}c@{}}300,000 - \\ 1,000,000\end{tabular} & \begin{tabular}[c]{@{}c@{}}150,000 -\\  299,000\end{tabular} & \begin{tabular}[c]{@{}c@{}}100,000 -\\  149,999\end{tabular} \\ \hline
\multicolumn{2}{|c|}{\begin{tabular}[c]{@{}c@{}}Total Square \\ Footage ($ft^2$)\end{tabular}} & 3830 & 2160 & 3130 \\ \hline
\end{tabular}
\vspace{-0.5cm}
\label{tb:infor}
% \end{table}
\end{table}

\begin{table*}[!btph]
\centering
\caption{The Average MSE Comparison between The Models}
\label{aver_MSE}
\begin{tabular}{|c|c|c|c|c|}
\hline \hline
% \begin{tabular}
% [c]{@{}c@{}}Analytic \\ Model\end{tabular}
{\multirow{2}{*}{\begin{tabular}[c]{@{}c@{}}Analytical Model\end{tabular}}} &
% \begin{tabular}[c]{@{}c@{}}Average\\ MSE\end{tabular} 
{\multirow{2}{*}{\begin{tabular}[c]{@{}c@{}}Average MSE\end{tabular}}}
& 
\multicolumn{3}{c|}{\begin{tabular}[c]{@{}c@{}} Error Reduction\\Compared with (\%)\end{tabular}}   \\ \cline{3-5} 
{}&{}&Benchmark1 &Benchmark2&Benchmark3\\ \cline{2-4} 
% \begin{tabular}[c]{@{}c@{}} Error Reduction\\Compared with\\ Benchmark 1 (\%)\end{tabular} & \begin{tabular}[c]{@{}c@{}}Error Reduction\\Compared with\\ Benchmark 2 (\%)\end{tabular}& \begin{tabular}[c]{@{}c@{}}Error Reduction\\Compared with\\ Benchmark 3 (\%)\end{tabular} \\ 
\hline \hline
\multicolumn{5}{|c|}{Weekday}
\\ \hline
{\begin{tabular}[c]{@{}c@{}}Benchmark 1\end{tabular}} & 0.134 & \cellcolor[HTML]{656565} & \cellcolor[HTML]{656565} & \cellcolor[HTML]{656565} \\ \hline
{\begin{tabular}[c]{@{}c@{}}Benchmark 2\end{tabular}} & 0.038 & 71.6 & \cellcolor[HTML]{656565} & \cellcolor[HTML]{656565}\\ \hline
{\begin{tabular}[c]{@{}c@{}}Benchmark 3\end{tabular}} & 0.022 &83.6& 42.5 & \cellcolor[HTML]{656565} \\ \hline
{\begin{tabular}[c]{@{}c@{}}Proposed Model\end{tabular}} & 0.017&87.3 & 54.2 & 20.4 \\ \hline
\multicolumn{5}{|c|}{Weekend} \\ \hline
{\begin{tabular}[c]{@{}c@{}}Benchmark 1\end{tabular}} & 0.072 & \cellcolor[HTML]{656565} & \cellcolor[HTML]{656565}& \cellcolor[HTML]{656565} \\ \hline
{\begin{tabular}[c]{@{}c@{}}Benchmark 2\end{tabular}} & 0.012 & 83.7& \cellcolor[HTML]{656565}& \cellcolor[HTML]{656565} \\ \hline
{\begin{tabular}[c]{@{}c@{}}Benchmark 3\end{tabular}} & 0.010 &86.5& 16.7 & \cellcolor[HTML]{656565} \\ \hline
{\begin{tabular}[c]{@{}c@{}}Proposed Model\end{tabular}} & 0.009&87.8& 23.0 & 7.0 \\ \hline
\end{tabular}
\end{table*}

\subsubsection{Overall Performance}
We also use MSEs in Equation (\ref{MSEeq}) to measure the errors of all compared methods, as shown in TABLE \ref{aver_MSE} for both weekdays and weekends.
% Fig. \ref{predict_fig} according to individual load patterns and dataset category. 
% % The prediction results gained by the testing on benchmark 1, benchmark 2, and proposed model are denoted by blue line, orange dash line, and green dash line separately. 
% We can see that both our model and benchmark 2 outperform the baseline method distinctly for all the load patterns besides G6 of weekday and G5 of weekend. To be more specific, the joint structure does improve the accuracy for the overall prediction. However, the enrichment brought by the feature selection is still visible as the prediction error is quite small comparing with the baseline. 
% \begin{figure}[tp!] 
% \centering\includegraphics[width=0.8\columnwidth]{pre_comparison.pdf}
% \vspace{-0.5cm}
% \caption{The performance evaluation through three prediction models.}\vspace{-0.7cm} \label{predict_fig} \end{figure}
Benchmark 1 does not perform well on both weekdays and weekends. Employing DNN, Benchmark 2 obtains a noticeable improvement compared with Benchmark 1. However, using our proposed DNN structure, both benchmark 3 and our model significantly reduce the errors by 42.5\% and 54.2\% compared with Benchmark 2. Using feature selection, our model achieves a further reduction of 20.4\% in errors compared to Benchmark 3. Moreover, as the $Cluster Score$ for the weekend load pattern is lower than that for the weekday load pattern, the overall MSE for weekends is lower than weekdays. Note that the performance in terms of errors will be affected by data, and we do not intend to emphasize on percentage improvement made by our model. Instead, the results demonstrate that our model better captures the nonlinearity between the load pattern and socioeconomic features.
%\vspace{-0.4cm}

% when applying to weekend load pattern. 
% The joint structure firstly bring a 17.1\% growth of prediction accuracy, proved by the comparison between benchmark 2 and baseline, while a 7\% growth can be deducted achieved by additionally involving feature selection. 

\section{Conclusion}
\label{sec:conclu}
% To study how energy consumption behaviors of residential consumers are correlated with % various factors, especially socioeconomic characteristics,
We developed an analytical method to advance the understanding of residential electricity load patterns by focusing on the impact of consumers' socioeconomic factors.
% especially in the situation of lacking the smart meter data,
%  using clustering and deep learning algorithm. 
% realistic data.
% that aims to undercover the residential load pattern using a realistic load dataset. The tool creatively leverages the ideas of data mining and deep learning. Specifically
% , the stepwise proceedings including data pre-processing, load pattern capturing, feature selection, and finally prediction model are implemented. W
Specifically, we used K-Medoid clustering to identify representative load patterns, given K-Medoid's advantage of being robust to outliers. We also used the entropy-based feature selection method to obtain the pattern-related feature sets. The feature selection contributed to identifying the critical socioeconomic factors on different load patterns and improving the interpretability of our method.
% according to each load pattern 
% which exclude the negative effects, e.g. high computing cost caused by redundancy, 
% for the following prediction
Then we developed a deep learning model to reveal the relationship between the distribution of load patterns and the selected socioeconomic features. Our model consists of pattern-dependent DNNs and a normalization layer to enhance the accuracy. Our model consists of pattern-dependent DNNs and a normalization layer to enhance the accuracy. We summarize the results based on the realistic load data and socioeconomic data as follows.
{
\begin{enumerate}
    \item We obtained $6$ representative daily load curves to model the consumption patterns of weekdays, and all $6$ load curves showed evening peaks. Also, $6$ daily load curves were selected to model the consumption on weekends, and a single peak or dual peaks during 9:00-18:00 were more commonly found.
    \item The age and education level were found to be two significant drivers of load patterns. 
    % The age of greater than $65$ exhibited a higher impact on the weekday load patterns, while the age ranging from 12 to 49 has a higher correlation with weekend load patterns. Moreover, 
    Meanwhile, the education level has a strong effect on all load patterns.
    \item It can be observed that consumers with age older than 65 make a great impact on the weekdays' typical consumption curves. The people aged older than 65 are thus potentially the target participants in demand response programs.
    \item The rich information can be revealed by the DNN model through analyzing how individual household consumes energy and how diverse socioeconomic factors make the effect on the consumption.
    \item The comparison results with other benchmark methods show the non-linear relationship between the selected features and load patterns. Our model is better at mapping the socioeconomic features to the load patterns with an average error reduction of 46.5\% compared to benchmark methods.
    % \item Numerical results, including the individual analysis and overall performance, validated our proposed analytical model in revealing the relationship between load patterns and socioeconomic factors.
\end{enumerate}
}

{
For our future work, 
% we will separate the data referring to diverse categories, e.g. season, holiday, or different weather, to have a deeper decomposition of the factors under the consumption. Furthermore, 
% data-driven planning and operation strategies, e.g., the recommendation of demand response program, Time of Use tariff plan to consumer selection, and long-term ISO dispatch planning, for the system operator based on the load analysis results in this work will be developed. 
we plan to}
\begin{enumerate}
   \item {study appliance consumption data and consider categories, e.g., season, holiday, or different weather;}
    \item {analyze the price elasticity of the consumption by considering price factors as the inputs of the DNN model;} 
    \item {and further develop energy programs, such as demand response, using the results of the consumers' load pattern probabilities.}
\end{enumerate}

% the application of our proposed model and related results will be studied, e.g. the recommendation of demand response program or Time of Use tariff plan to consumer selection.}

% In this work, we had discussed the relation between the consumption behavior and socioeconomic information, which is the household endogenous factors. 
% Our future work aims to develop data-driven planning and operation strategies for the system operator based on the load analysis results in this work.
%, especially when some users' meter readings are not available.
\vspace{-0.5cm}
% the  the research on the impact made by the endogenous and external factors, e.g. weather, date, season, etc., jointly on the consumption behavior will be further studied. 
% Therefore, we plan to have the deeper decomposition of the relations between these factors and consumption behavior, which, we believe, can further benefit for the system operation.
% This work  The consumption behavior For future work, we will categorize the consumption data referring to season, holiday, and weather condition, to have a deeper decomposition of the socioeconomic factors under the consumption.  we aim to study the transient effect between each load pattern, based on which the analysis between transient effect and socioeconomic information will be made.

%\section*{Acknowledge}
%This work was supported in part by National Applied Research Laboratories (108-3116-F-492-002-MY2), Taiwan.

%% The Appendices part is started with the command \appendix;
%% appendix sections are then done as normal sections
%% \appendix

%% \section{}
%% \label{}

%% If you have bibdatabase file and want bibtex to generate the
%% bibitems, please use
%%
\bibliographystyle{elsarticle-num-names} 
% \bibliography{<ref>}

%% else use the following coding to input the bibitems directly in the
%% TeX file.

% \begin{thebibliography}{ref}

% %% \bibitem[Author(year)]{label}
% %% Text of bibliographic item

% \bibitem[ ()]{}
\bibliography{ref.bib}

%% else use the following coding to input the bibitems directly in the
%% TeX file.

% \begin{thebibliography}{00}

% %% \bibitem[Author(year)]{label}
% %% Text of bibliographic item

% \bibitem[ ()]{}

% \end{thebibliography}
\end{document}